\documentclass[10pt,twocolumn,letterpaper]{article}
\usepackage[T1]{fontenc}
\usepackage[pagenumbers]{cvpr}
\usepackage{graphicx,amsmath,amssymb,booktabs,array,multirow,xcolor,colortbl,tabularx,float,needspace}
\definecolor{cvprblue}{rgb}{0.21,0.49,0.74}
\usepackage[pagebackref,breaklinks,colorlinks,allcolors=cvprblue]{hyperref}
\DeclareRobustCommand{\docref}[2]{\ref{#1}}
\def\confName{CVPR}\def\confYear{2027}
\newcommand{\method}{GeoScope\xspace}
\newcommand{\bench}{GeoScope-Bench\xspace}
\newcommand{\masktoken}{\texttt{<|mask|>}}
\title{GeoScope: What to Mask and What to Merge in Remote Sensing Reasoning Segmentation}
\author{%
  Yichang Luo\textsuperscript{1,2},
  Shixiong Liu\textsuperscript{1$\dagger$},
  Kan Wei\textsuperscript{1,2},
  Yan Zhao\textsuperscript{1,2},
  Boxiang Yang\textsuperscript{3},
  Lei Wang\textsuperscript{1,2}\\[4pt]
  \textsuperscript{1}Aerospace Information Research Institute, Chinese Academy of Sciences\\
  \textsuperscript{2}University of Chinese Academy of Sciences\quad
  \textsuperscript{3}Peking University
}
\begin{document}
\maketitle
\begingroup
\renewcommand\thefootnote{\fnsymbol{footnote}}
\footnotetext[2]{Corresponding author.}
\endgroup

\setlength{\textfloatsep}{10pt plus 2pt minus 2pt}
\setlength{\floatsep}{8pt plus 2pt minus 2pt}
\setlength{\intextsep}{8pt plus 2pt minus 2pt}

\raggedbottom

\begin{abstract}
Remote sensing reasoning segmentation requires models to infer requested regions, delineate their spatial extent, and organize masks into specified output groups. Visually similar instances often induce duplicate localization, and accurate foreground coverage can still accompany incorrect grouping. Existing benchmarks evaluate queries in isolation, offering limited insight into whether predictions track changes in target selection and grouping on the same scene. Thus, we construct \bench with 851 same-image instruction pairs annotated with reference masks and output groups across five controlled transitions: identity, extent, membership, partition, and paraphrasing. We further propose \method to explicitly decouple atomic target localization from topological group assembly via a structured plan. To prevent similarly described targets from collapsing onto one instance, Joint Mask-Token Readout concatenates target segments in a single causal sequence, conditioning later mask-token states on earlier target context for implicit feature-space de-duplication. Experiments show that \method substantially leads on \bench and achieves the highest mean scores on EarthReason and three zero-shot referring benchmarks.
The benchmark and code will be publicly released at
\href{https://github.com/YeeCharm/GeoScope}{\textcolor{cvprblue}{\textit{https://github.com/YeeCharm/GeoScope}}}.
\end{abstract}

\section{Introduction}
\label{sec:intro}
\begin{figure}[t]
\centering
\includegraphics[width=\linewidth]{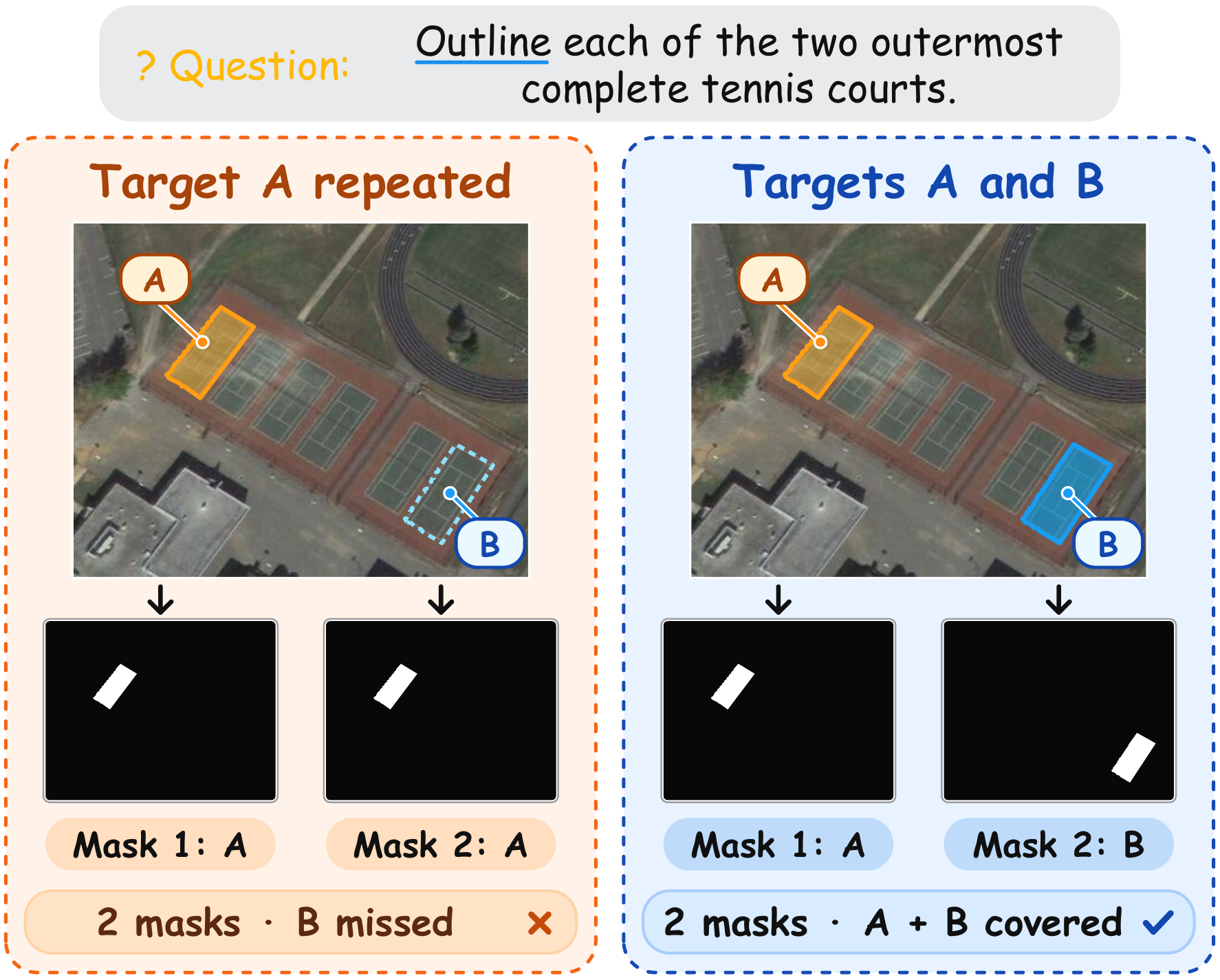}
\captionsetup{skip=4pt}
\caption{\textbf{Two masks can still miss a target.} The request asks for a separate mask for each outermost complete tennis court. Both responses return two masks. The left response localizes court A twice, leaving court B uncovered.}
\label{fig:motivation}
\end{figure}

Remote sensing reasoning segmentation grounds natural-language instructions in aerial images by inferring the requested regions from semantic and spatial cues and producing their pixel masks~\cite{segearthr1,segearthr2}. Relative positions and nearby landmarks help distinguish similar objects in a scene~\cite{zoomearth,ni2026unigeoseg,luo2026cross}. Accurate execution requires identifying the intended targets and their number~\cite{kim2026improving}, delineating their extent, and grouping masks as requested. Duplicate localization leaves a target uncovered even at the requested mask count (e.g., two masks on one court in Figure~\ref{fig:motivation}). 

For requests to merge and to separate the same targets, the expected foreground stays fixed and the output groups differ. Existing benchmarks compare each response with its reference masks across a range of queries and target granularities~\cite{segearthr1,segearthr2}. Testing a change in the request also requires a reference for how the two outputs should relate. Following controlled grounding studies~\cite{zeng2024investigating,li2026counterfactual}, we pair requests on the same image to test which regions and groups change or stay fixed.

To measure execution under such changes, we introduce \bench with 851 request pairs on 404 aerial images (Figure~\ref{fig:bench}). Both requests use the same image and have reference target masks and output groups. The pairs cover changes in target identity, extent, membership, and partition, along with consistency under paraphrasing. When the second request adds a target, its output should retain the first mask and add one for the new target. Models process each request separately. Pair Success is the share of pairs whose outputs meet both requests and the specified relation.

Executing such scope-sensitive requests exposes a key bottleneck in existing MLLM segmenters: a single \texttt{<SEG>} token must simultaneously resolve instance-level spatial disambiguation and group-level mask topology. We propose \method to explicitly decouple \textit{what to localize} from \textit{how to merge}. Structured Target and Group Planning first decomposes an instruction into indexed atomic targets and symbolic output groups, allowing the same localized instances to be merged or separated without re-grounding.

Furthermore, when multiple targets share identical category descriptions, independent readout per target produces homogeneous queries that collapse onto the same salient instance. To resolve this ambiguity directly during sequence modeling, we introduce Joint Mask-Token Readout for implicit feature-space de-duplication. By concatenating all planned target segments with a designated \masktoken{} token after the shared image and instruction prefix, causal self-attention exposes later mask-token states to earlier target representations, shifting their spatial queries toward uncovered instances before decoding and group union assembly.

Our main contributions are summarized as follows:
\begin{itemize}
\item We construct \bench (851 request pairs, 1,692 distinct instructions) to evaluate scope-aware instruction execution across controlled same-image transitions in target identity, extent, membership, partition, and paraphrasing.
\item We propose \method, which decouples atomic target planning from topological group assembly and introduces causal Joint Mask-Token Readout to resolve multi-instance query homogeneity in latent space.
\item Extensive experiments show that \method achieves the highest Pair Success on \bench and leads public reasoning and zero-shot referring segmentation benchmarks, with pronounced gains in target addition and regrouping.
\end{itemize}

\section{Related Work}
\label{sec:related}

\subsection{Grounded Visual Reasoning}
\label{sec:grounded_related}
Grounded visual reasoning links language-based inference to localized evidence in an image. Multimodal segmenters delineate this evidence with pixel masks~\cite{flmm,read,mllmseg}, and region masks provide context for subsequent reasoning or dialogue~\cite{wang2025segllm,liu2026unipixel,terrascope}. LISA~\cite{lisa} grounds implicit instructions by decoding the hidden state of a \texttt{<SEG>} token into a mask. PixelLM~\cite{pixellm} uses a segmentation codebook and a lightweight decoder for multiple targets, with a refinement loss to reduce overlap between their masks. Multi-target grounding also requires binding each description to its intended instance, a challenge in scenes containing several visually similar objects. The spatial relations specified in the request constrain which combinations of instances are valid, even when the predicted masks are disjoint.

\subsection{Remote Sensing Reasoning Segmentation}
Remote sensing referring segmentation grounds explicit expressions~\cite{refsegrs,rmsin,risbench,chen2025rsrefseg}, and generalist systems cover dialogue, box localization, and masks~\cite{geochat,geopixel,zhou2024geoground,ou2025geopix,li2024vrsbench}. SegEarth-R2~\cite{segearthr2} combines spatial attention supervision for small objects and parts with flexible queries for single and multiple targets. Think2Seg-RS~\cite{think2segrs} prompts a frozen SAM2 with boxes and points predicted by a vision-language model. Related work expands instruction coverage~\cite{ni2026unigeoseg,zheng2026instructsam} or refines localization through structural priors, active perception, and finer spatial representations~\cite{georeason,zoomearth,pixdlm,firm2026}. In dense aerial scenes, spatial qualifiers narrow the candidate set. Part-level grounding further depends on local boundaries and the location of the component within its parent structure.

\subsection{Target Selection and Mask Grouping}
Phrase grounding links language expressions to regions at the pixel level~\cite{wu2020phrasecut,glamm,pixellm}. GRES~\cite{gres} and related methods extend referring segmentation to requests with multiple or no targets~\cite{gsva}, with further work addressing part-level and compositional requests~\cite{wang2024unveiling,jang2025mmr,groundv,cai2025naver}. RAS~\cite{cao2025refer} starts from masks proposed by a segmenter and selects a group based on text and optional visual references. The same foreground admits different output groupings depending on the instruction (\textit{e.g.}, returning two storage tanks as a single merged mask or as two separate masks). For three selected targets, $\{A,B\},\{C\}$ and $\{A\},\{B,C\}$ yield the same mask count with different assignments to output groups.

\section{GeoScope-Bench}
\label{sec:bench}

\subsection{Data Design}
\begin{figure*}[t]
\centering
\includegraphics[width=\textwidth]{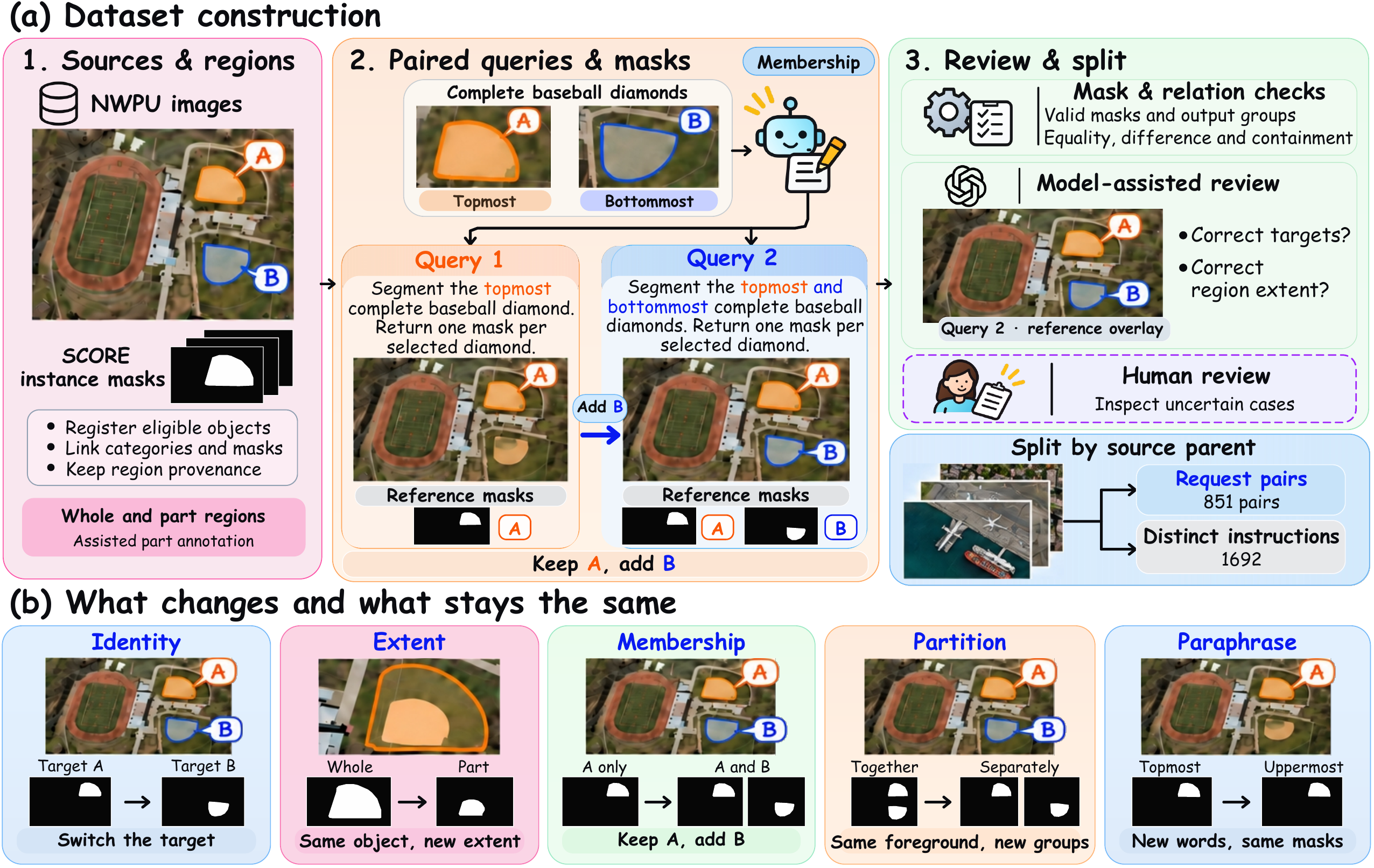}
\captionsetup{skip=4pt}
\caption{\textbf{GeoScope-Bench construction and evaluation pairs.} \textbf{(a)} We register source regions, construct paired requests with reference masks, and check the annotations before splitting by source parent. \textbf{(b)} Identity, extent, membership and partition each change a different aspect of the requested output. Paraphrasing preserves the reference masks and their grouping.}
\label{fig:bench}
\end{figure*}

Existing remote sensing segmentation benchmarks~\cite{segearthr1,segearthr2} assess each prediction against the reference masks for its instruction. Evaluating the effect of an instruction change requires paired requests on the same image and a reference for how the outputs should relate. We construct \bench through the three stages in Figure~\ref{fig:bench}, pairing requests with annotated target masks and output groups.

\subsection{Data Construction and Quality Control}
\noindent\textbf{Stage 1. Sources and Regions.} To strictly isolate scope transitions under controlled multi-instance topology, \bench curates multi-object aerial scenes from NWPU-VHR-10~\cite{su2019precise} with fine-grained instance masks from SCORE~\cite{score}, ensuring co-occurring instances of the same category per scene. Each registered region retains its category, binary mask, and provenance. Deterministic spatial ranking identifies unambiguous positional targets (\textit{e.g.}, outermost or southernmost instances), complemented by verified part polygons for whole-to-part revisions.

\noindent\textbf{Stage 2. Paired Queries and Evidence.} The registered regions provide reference masks for instruction pairs on the same image, with each pair isolating one of five transformations (Figure~\ref{fig:bench}(b)).
\begin{itemize}
\setlength{\itemsep}{0pt}
\setlength{\topsep}{2pt}
\setlength{\parsep}{0pt}
\item \textbf{Identity.} Parallel spatial qualifiers select different instances of one category (\textit{e.g.}, the topmost or bottommost complete baseball diamond). The reference switches instances at the same granularity and mask count.
\item \textbf{Extent.} A whole-to-part revision changes the requested extent and preserves the selected instance and output count. The part-level reference delineates the specified subregion within the whole-object mask.
\item \textbf{Membership.} The second request adds an eligible instance under the same per-instance output rule. Its reference retains the original mask and adds a separate mask for the new instance (Figure~\ref{fig:bench}(a)).
\item \textbf{Partition.} The two requests share the selected regions and their foreground union. One reference merges these regions into a single mask, and the other assigns each instance to a separate output mask.
\item \textbf{Paraphrase.} A semantically equivalent revision changes the wording (\textit{e.g.}, replacing \textit{topmost} with \textit{uppermost}). Both requests share the same targets, spatial extents, and output groups.
\end{itemize}

\noindent\textbf{Stage 3. Review and Verification.} Programmatic checks enforce valid mask dimensions, nonempty foregrounds, exact group counts, mask disjointness for target switches, union equality for partitions, and spatial containment for whole--part pairs. After assisting candidate query generation, GPT-6.1 Sol~\cite{openai2026gpt61sol} pre-screens the image, paired queries, and reference overlays to flag ambiguities, followed by human verification and manual auditing to eliminate self-bias. Split strictly by source image to prevent leakage, \bench comprises 404 images, 851 pairs, and 1,692 distinct instructions across the five transformations (audit details in Appendix~\docref{app:construction}{A}).

\subsection{Evaluation Protocol}
Models process the two requests on the same image independently. Union scores compare the predicted and reference foreground unions. Group scores use one-to-one Hungarian assignment~\cite{kuhn1955hungarian} to match masks and account for unmatched groups. For both Union and Group, gIoU averages per-request overlaps and cIoU pools intersections and unions. Merging two correctly delineated instance masks preserves Union IoU and lowers Group IoU when the request specifies separate masks.

Request Pass measures the percentage of requests satisfying check $A(I,q)$, which requires exact group count $K$, matched group $\text{IoU}\ge 0.5$, and correct target binding. Pair Success requires both requests to pass $A(\cdot)$ and satisfy the transformation-specific relation $C_i$ (formal definitions and threshold sensitivity in Appendix~\docref{app:metrics}{B}). For $N$ pairs,
\begin{equation}
 \mathrm{Pair}=\frac{100}{N}\sum_{i=1}^{N}
   \mathbf{1}\{A(I_i,q_i)\land A(I_i,q'_i)\land C_i\}.
 \label{eq:pair}
\end{equation}

\section{GeoScope}
\label{sec:method}
\begin{figure*}[t]
\centering
\includegraphics[width=\textwidth]{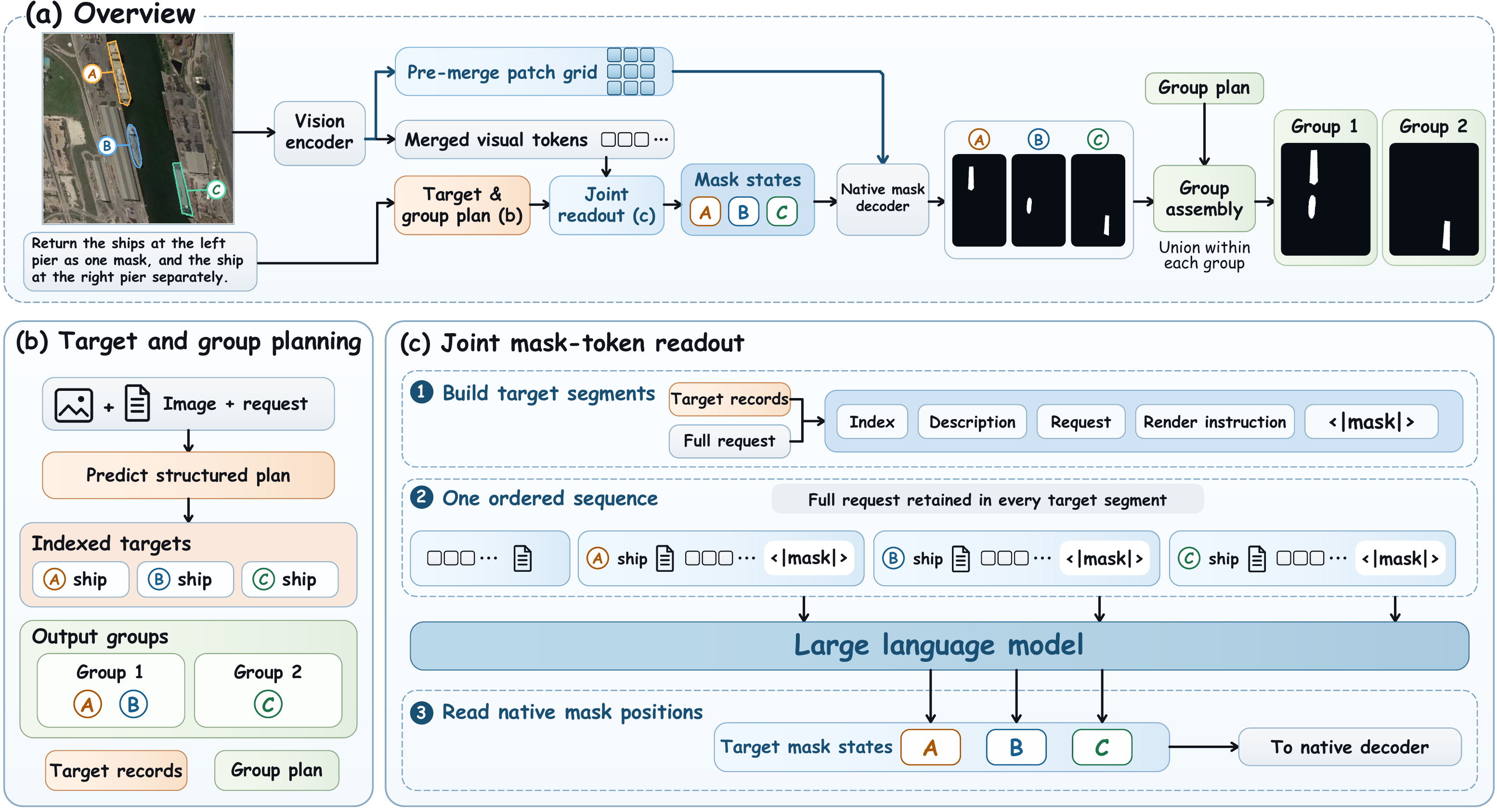}
\captionsetup{skip=4pt}
\caption{\textbf{GeoScope overview.} \textbf{(a)} Target states are decoded against the shared pre-merge patch grid and combined within each output group. \textbf{(b)} The plan indexes the ships and places the two at the left pier in a common group. \textbf{(c)} Joint readout concatenates indexed target segments and extracts the hidden state at each mask-token position.}
\label{fig:method}
\end{figure*}

Building on the observation that duplicate localization occurs even at the requested mask count (Figure~\ref{fig:motivation}), \method specifies target regions and output groups in an indexed plan. Joint readout uses the plan order to condition later mask-token states on preceding target segments. As shown in Figure~\ref{fig:method}, each indexed state is decoded into a spatial mask, and the plan determines how these masks are assembled into the requested outputs.

\begingroup
\setlength{\abovedisplayskip}{4pt plus 1pt minus 1pt}
\setlength{\belowdisplayskip}{4pt plus 1pt minus 1pt}
\setlength{\abovedisplayshortskip}{2pt plus 1pt}
\setlength{\belowdisplayshortskip}{3pt plus 1pt minus 1pt}
\setlength{\jot}{2pt}
\subsection{Structured Target and Group Planning}
An instruction specifies both the regions to render and their assignment to output masks. Given an aerial image $I\in\mathbb{R}^{3\times H\times W}$ and instruction $q$, the planner generates $A$ indexed descriptions and $K$ output groups,
\begin{equation}
 P(I,q)=\big((d_i)_{i=1}^{A},(G_k)_{k=1}^{K}\big).
 \label{eq:plan}
\end{equation}

Each $d_i$ describes a target region (\textit{e.g.}, an instance, a part, or a set of objects). Its place in the ordered list provides a persistent index even when several targets share a category description. Each nonempty $G_k\subseteq\{1,\ldots,A\}$ specifies which target indices contribute to output mask $k$.

The planner serializes the descriptions and groups as separate lists. Group assignments refer to the target indices even when the descriptions coincide (\textit{e.g.}, the three targets described as \textit{ship} in Figure~\ref{fig:method}(b) form $G_1=\{1,2\}$ for the left pier and $G_2=\{3\}$ for the right pier). The indexed descriptions then define the target segments in plan order.

\subsection{Joint Mask-Token Readout}
Short descriptions of similar objects often omit the spatial qualifiers that identify an instance. As shown in Figure~\ref{fig:method}(c), each target segment $c_i$ retains the full request $q$ alongside the target index, planned description $d_i$ (preserving explicit spatial clauses when present or defaulting to $q$ for single-target queries), and a rendering directive with one designated \masktoken{} position.

Both readouts use the same target segments and image and instruction prefix $\pi(I,q)$. Let $\Vert$ denote concatenation. Independent readout appends one segment to the prefix per forward pass. Joint readout appends all segments in plan order for a single forward pass,
\begin{equation}
 \begin{aligned}
 S^{\mathrm{joint}} &= \pi(I,q)\mathbin{\Vert}c_1\mathbin{\Vert}\cdots\mathbin{\Vert}c_A,\\
 S_i^{\mathrm{ind}} &= \pi(I,q)\mathbin{\Vert}c_i .
 \end{aligned}
 \label{eq:sequence}
\end{equation}

Causal attention restricts each state to tokens at or before its designated mask-token position. Both readouts expose the shared prefix and the portion of $c_i$ through that position. The sequence of preceding target segments $E_i$ depends on the readout,
\begin{equation}
 E_i^{\mathrm{joint}}=c_1\mathbin{\Vert}\cdots\mathbin{\Vert}c_{i-1},
 \qquad E_i^{\mathrm{ind}}=\varepsilon.
 \label{eq:causal_attn}
\end{equation}

Here $\varepsilon$ denotes the empty sequence. The first target has empty preceding target context in both readouts, and the two inputs coincide for $A=1$. In Figure~\ref{fig:method}(c), the independent state for ship B is conditioned on the shared prefix and its own segment $c_2$ through the mask token. Under joint readout, that state also has access to the tokens of ship A's segment $c_1$, including the earlier mask token.

The renderer extracts the final-layer state for target $i$ at its designated mask-token position $m_i$ in the joint sequence,
\begin{equation}
 h_i^{\mathrm{joint}}=\big[\operatorname{LM}_{\theta_{\mathrm{r}}}(S^{\mathrm{joint}})\big]_{m_i}.
 \label{eq:joint_state}
\end{equation}

Independent readout extracts the state at the corresponding position in $S_i^{\mathrm{ind}}$ using the same renderer. When multiple targets share a descriptor $d_i$ (\textit{e.g.}, \textit{ship}), independent passes receive identical prefix and lexical cues, yielding nearly collinear queries $w_i$ in Eq.~\eqref{eq:dot_prod} that collapse onto the same dominant instance. Conversely, Joint Readout exposes $h_i^{\mathrm{joint}}$ to $E_i^{\mathrm{joint}}$, including earlier \masktoken{} states already bound to specific visual regions. Through causal self-attention, the LLM implicitly shifts $h_i^{\mathrm{joint}}$ away from already-attended instances in latent space prior to projection, steering $w_i$ toward remaining valid candidates on the shared grid $F(x)$.

The mask-token position $m_i$ locates the state in the sequence, and the target index $i$ links that state to its planned output groups. Accordingly, the order control reverses the segments and maps the states back to their original indices before assembly. The rotary position control changes the position IDs at a fixed segment order. Appendix~\docref{app:implementation}{C} specifies the segment template shared by both readouts.

\subsection{Mask Decoding and Group Assembly}
To localize the indexed targets, the mask decoder scores a spatial grid shared by all readout states. The two visual paths in Figure~\ref{fig:method}(a) originate from the encoder's pre-merge patch grid $V^0$, which supplies merged tokens $V$ for the language-model prefix and spatial embeddings $F$ for mask decoding,
\begin{equation}
 (V,F)=\big(\operatorname{Merge}(V^0),\mathcal D_{\mathrm v}(V^0)\big).
 \label{eq:grids}
\end{equation}

The feature transform $\mathcal D_{\mathrm v}$ produces $F$ with channel dimension $d_m$. A two-layer MLP $\mathcal P_m$ projects each state $h_i$ into this space. The decoder scores position $x$ by matching $F(x)$ to the scaled target query $w_i$,
\begin{equation}
 w_i=d_m^{-1/2}\mathcal P_m(h_i),\qquad
 Z_i(x)=\big\langle F(x),w_i\big\rangle .
 \label{eq:dot_prod}
\end{equation}

Let $\mathcal U_I$ resize each logit map to the output resolution and $\sigma$ denote the sigmoid. Thresholding the resulting probabilities yields one binary mask per target. Each output mask is the union of the target masks assigned to its group in Eq.~\eqref{eq:plan},
\begin{equation}
 \begin{aligned}
 \widehat A_i&=\mathbf{1}\!\left[\sigma\!\left(\mathcal U_I(Z_i)\right)\geq 0.5\right],\\
 \widehat M_k&=\bigvee_{i\in G_k}\widehat A_i.
 \end{aligned}
 \label{eq:assembly}
\end{equation}

A single output mask may thus comprise disjoint regions (\textit{e.g.}, $G_1$ in Figure~\ref{fig:method}(b) combines the two ships at the left pier). Given fixed target masks, regrouping preserves the foreground union whenever the groups cover all targets.

\subsection{Planner and Renderer Adaptation}
The planner and renderer are adapted separately and share a frozen vision encoder. The planner learns the serialized target and group plan by training its language LoRA~\cite{hu2022lora} with token cross-entropy. The renderer's language LoRA and native mask decoder are adapted with token and pixel supervision. Given $B$ supervised masks, $\bar Z_i$ denotes the logits on the decoder canvas, and $Y_i$ denotes the corresponding reference mask resized to that canvas by nearest-neighbor interpolation. The per-mask loss $\ell_{\mathrm{mask}}(z,y)$ adds pixelwise binary cross-entropy on $z$ and Dice loss on $\sigma(z)$, both evaluated against $y$. The rendering objective combines this term with next-token cross-entropy $\mathcal L_{\mathrm{tok}}$,
\begin{equation}
 \mathcal L_{\mathrm r}
 =\mathcal L_{\mathrm{tok}}+\frac{1}{B}\sum_{i=1}^{B}
   \ell_{\mathrm{mask}}(\bar Z_i,Y_i).
 \label{eq:renderer_loss}
\end{equation}

Mask supervision acts on the decoder-canvas predictions before thresholding and assembly. Because Joint Readout processes all $A$ segments in a single pass over the shared prefix $\pi(I,q)$ without recomputing visual tokens, it incurs negligible memory overhead ($<5\%$) and lower latency than $A$ independent passes (runtime scaling in Appendix~\docref{app:implementation}{C}).
\endgroup

\section{Experiments}
\label{sec:experiments}
\raggedbottom

\begin{table*}[t]
\centering\footnotesize
\captionsetup{skip=5pt}
\caption{\textbf{Evaluation on GeoScope-Bench (\%).} The five transformation columns report Group gIoU. The final two columns give overall Group gIoU/cIoU and Pair Success. Bold and underlining indicate the highest and second-highest scores, respectively.}
\label{tab:bench}
\setlength{\tabcolsep}{3.2pt}
\renewcommand{\arraystretch}{1.00}
\begin{tabular*}{\textwidth}{@{\extracolsep{\fill}}l*{7}{c}@{}}
\toprule
Method & Identity & Extent & Membership & Partition & Paraphrase & Group gIoU / cIoU & Pair Success\\
\midrule
LISA+-7B~\cite{yang2025lisaplus} & 30.48 & 46.60 & 23.57 & 29.56 & 36.09 & 32.03 / 33.44 & 10.00\\
GSVA-7B~\cite{gsva} & 16.73 & 26.99 & 12.42 & 11.01 & 15.99 & 15.66 / 15.34 & 2.00\\
PSALM~\cite{zhang2025psalm} & 13.97 & 32.86 & 7.81 & 8.91 & 14.74 & 14.07 / \phantom{0}7.87 & 3.33\\
PixelLM-7B~\cite{pixellm} & 31.95 & 20.78 & 31.25 & 20.88 & 29.13 & 27.27 / 24.81 & 11.33\\
GeoPixel-7B~\cite{geopixel} & 16.44 & 13.03 & 14.08 & 16.42 & 15.02 & 15.20 / 21.72 & 1.33\\
SegEarth-R1~\cite{segearthr1} & 13.15 & 40.99 & 11.05 & 16.00 & 29.85 & 20.44 / 22.37 & 4.67\\
Seg-R1-7B~\cite{you2025segr1} & 47.61 & 33.95 & 30.09 & 28.73 & 51.02 & 38.59 / 29.77 & 16.67\\
MLLMSeg-4B~\cite{mllmseg} & 41.51 & 51.57 & 26.50 & 25.35 & 53.40 & 38.46 / \underline{50.05} & \underline{18.67}\\
Think2Seg-RS-3B~\cite{think2segrs} & 46.23 & 42.36 & \underline{36.77} & 34.31 & 47.05 & 41.19 / 38.92 & 14.67\\
VisionReasoner-7B~\cite{liu2026visionreasoner} & \underline{48.00} & \underline{53.83} & 31.61 & \underline{35.64} & \underline{57.64} & \underline{44.50} / 44.49 & 16.00\\
RemoteReasoner-7B~\cite{yao2026remotereasoner} & 37.52 & 47.88 & 26.23 & 30.60 & 52.16 & 37.98 / 42.27 & 17.33\\
\midrule
GeoScope-4B (Ours) & \textbf{66.90} & \textbf{58.35} & \textbf{58.67} & \textbf{60.94} & \textbf{62.88} & \textbf{61.85} / \textbf{60.46} & \textbf{49.33}\\
\bottomrule
\end{tabular*}
\end{table*}

\begin{figure*}[t]
\centering
\includegraphics[width=\textwidth]{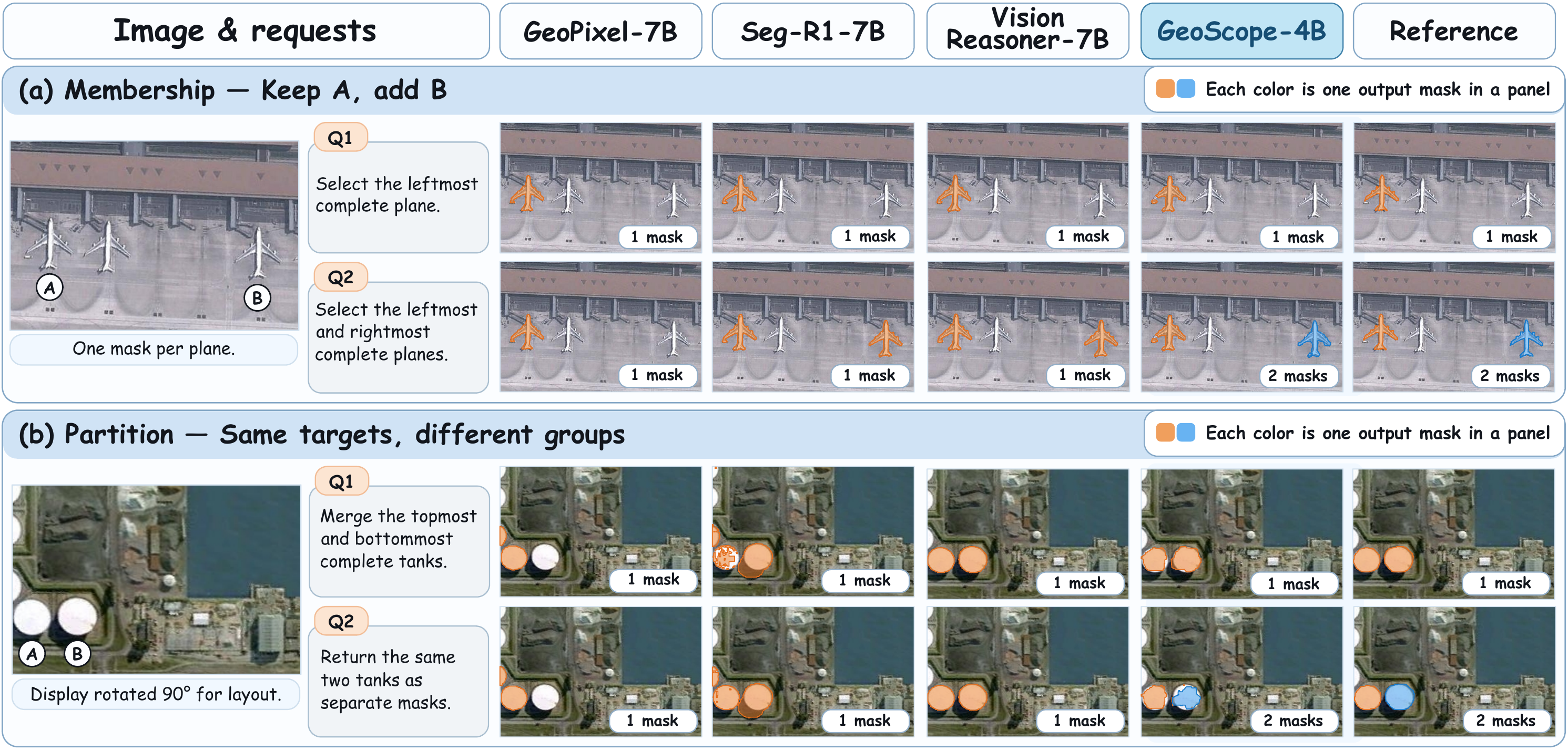}
\caption{\textbf{Selecting and grouping targets on GeoScope-Bench.} (a) A membership change adds the rightmost airplane as a separate mask. (b) A partition change keeps the two tanks and changes how their masks are returned. Colors identify individual output masks in each panel. We show saved development predictions.}
\label{fig:qualitative}
\end{figure*}

\subsection{Experimental Setup}
\noindent\textbf{Implementation.} We build \method on a Qwen3.5-4B~\cite{qwen2026qwen35} backbone. The vision encoder remains frozen, and planner adaptation uses language-model LoRA~\cite{hu2022lora}. For the renderer, we train the language-model LoRA and the mask decoder. The renderer's LoRA uses rank 64, $\alpha=128$, and dropout 0.05. Renderer adaptation uses BF16 forward computation and an effective batch size of 64, with peak learning rates of $2\times10^{-4}$ for LoRA and $2\times10^{-5}$ for the mask decoder. Appendix~\docref{app:implementation}{C} details the training schedule and inference settings.

\noindent\textbf{Benchmarks and metrics.} On \bench, we report Union and Group gIoU/cIoU, Request Pass, and Pair Success (Eq.~\eqref{eq:pair}), evaluating all methods under unified prompts and Hungarian group matching (adaptation controls in Appendix~\docref{app:implementation}{C}). For reasoning segmentation, we adapt on EarthReason~\cite{segearthr1} train and evaluate on val/test. For zero-shot referring segmentation, we directly evaluate on RRSIS-D~\cite{rmsin}, RISBench~\cite{risbench}, and RefSegRS~\cite{refsegrs}.

\subsection{Main Results}
\label{sec:bench_results}
As shown in Table~\ref{tab:bench}, we evaluate scope-aware instruction execution across five instruction transformations on the \bench test split.

\noindent\textbf{Identity.} \method leads in Group gIoU on Identity, where spatial qualifiers distinguish instances of the same category. A precise mask of the topmost diamond remains an error for a request specifying the bottommost one (Figure~\ref{fig:bench}(b)).

\noindent\textbf{Extent.} Extent yields the lowest Group gIoU for \method and the narrowest margin over the strongest baseline. A whole-object prediction may overlap the requested part, with pixels beyond the specified boundary lowering the score.

\noindent\textbf{Membership.} When the second request adds the rightmost airplane in Figure~\ref{fig:qualitative}(a), several baselines still return a single mask. \method retains the original selection and assigns a separate mask to the added instance.

\noindent\textbf{Partition.} Figure~\ref{fig:qualitative}(b) illustrates how a merged mask covers both tanks. Reusing it for the separate-mask request preserves the foreground union, with the incorrect group assignment lowering Group gIoU. \method covers both tanks in one mask for the merged request and in two masks for the separate request.

\noindent\textbf{Paraphrase.} Most baselines attain higher Group gIoU on Paraphrase than on Membership and Partition, which require changes to the target set or grouping.

\noindent\textbf{Overall performance.} \method attains 61.85\% Group gIoU and 49.33\% Pair Success. Its largest Group gIoU gains over the displayed baselines occur on Membership and Partition. Both changes depend on covering all specified instances in the output groups. The fixed-plan comparison in Section~\ref{sec:controlled} examines whether joint readout improves localization of later targets.

\subsection{Cross-Dataset Evaluation}
\label{sec:public_eval}
To assess \method beyond the paired requests in \bench, we evaluate two complementary remote sensing segmentation tasks. EarthReason tests localization from implicit requests after task adaptation. The zero-shot referring benchmarks assess localization from explicit descriptions across datasets.

\noindent\textbf{RS Reasoning Segmentation.} On EarthReason~\cite{segearthr1}, implicit requests identify targets through functional attributes or spatial relations rather than explicit category names. As shown in Table~\ref{tab:earthreason}, \method{} achieves the highest gIoU and cIoU across both validation and test splits, surpassing the strongest 4B and 7B baselines by +2.00 points on average.

\begin{table}[t]
\centering\footnotesize
\renewcommand{\arraystretch}{0.96}
\captionsetup{skip=4pt}
\caption{\textbf{Remote sensing reasoning segmentation on EarthReason (\%).} Avg.\ is the mean of the four validation and test scores. Bold and underlining denote the best and second-best scores in each column, respectively.}
\label{tab:earthreason}
\setlength{\tabcolsep}{2.8pt}
\renewcommand{\arraystretch}{0.96}
\begin{tabular*}{\columnwidth}{@{\extracolsep{\fill}}l*{5}{r}@{}}
\toprule
\multirow{2}{*}{Method} & \multicolumn{2}{c}{Validation} & \multicolumn{2}{c}{Test} & \multirow{2}{*}{Avg.}\\
\cmidrule(lr){2-3}\cmidrule(lr){4-5}
 & gIoU & cIoU & gIoU & cIoU & \\
\midrule
LISA~\cite{lisa} & 61.00 & 57.40 & 60.90 & 59.10 & 59.60\\
Seg-Zero~\cite{liu2025segzero} & 67.20 & 65.40 & 67.90 & 62.40 & 65.73\\
VisionReasoner~\cite{liu2026visionreasoner} & 66.60 & 67.10 & 67.20 & 65.40 & 66.58\\
RemoteReasoner~\cite{yao2026remotereasoner} & 69.00 & 67.80 & 71.00 & 69.10 & 69.23\\
SegEarth-R1~\cite{segearthr1} & 70.65 & 62.48 & 70.36 & 65.04 & 67.13\\
Text4Seg++~\cite{lan2026text4segpp} & 71.90 & 69.80 & 73.00 & 65.60 & 70.08\\
SegEarth-R2~\cite{segearthr2} & 72.30 & 68.10 & 73.50 & 69.50 & 70.85\\
UniGeoSeg~\cite{ni2026unigeoseg} & 72.54 & 73.30 & 73.08 & 74.61 & 73.38\\
Think2Seg-RS-7B~\cite{think2segrs} & 72.16 & 72.87 & 73.36 & 75.60 & 73.50\\
FIRM-4B~\cite{firm2026} & \underline{76.50} & \underline{77.50} & \underline{76.20} & \underline{75.90} & \underline{76.53}\\
\midrule
GeoScope-4B (Ours) & \textbf{77.10} & \textbf{78.82} & \textbf{77.76} & \textbf{80.43} & \textbf{78.53}\\
\bottomrule
\end{tabular*}
\end{table}

\noindent\textbf{RS Referring Segmentation.} Despite using a compact 4B backbone against mostly 7B baselines, \method substantially leads on RRSIS-D (+11.78 gIoU) and RISBench (+13.95 gIoU), driving the highest three-benchmark average (+9.09 points over VisionReasoner-7B in Table~\ref{tab:referring}). It trails slightly on RefSegRS~\cite{refsegrs}, where tiny dispersed targets ($<5\%$ image area) make IoU highly sensitive to minor boundary shifts and missing small instances.

\begin{table}[t]
\centering\footnotesize
\renewcommand{\arraystretch}{0.96}
\captionsetup{skip=4pt}
\caption{\textbf{Zero-shot remote sensing referring segmentation on the test splits (\%).} Avg.\ is the mean of the six gIoU and cIoU scores. Bold and underlining denote the best and second-best scores in each column, respectively.}
\label{tab:referring}
\setlength{\tabcolsep}{1.3pt}
\begin{tabular*}{\columnwidth}{@{\extracolsep{\fill}}l*{7}{c}@{}}
\toprule
\multirow{2}{*}{Method} & \multicolumn{2}{c}{RRSIS-D} & \multicolumn{2}{c}{RISBench} & \multicolumn{2}{c}{RefSegRS} & \multirow{2}{*}{Avg.}\\
\cmidrule(lr){2-3}\cmidrule(lr){4-5}\cmidrule(lr){6-7}
 & gIoU & cIoU & gIoU & cIoU & gIoU & cIoU & \\
\midrule
GSVA-7B~\cite{gsva} & 27.44 & 32.62 & 26.26 & 29.76 & \phantom{0}8.83 & 17.85 & 23.79\\
PixelLM-7B~\cite{pixellm} & 30.61 & 32.36 & 31.51 & 26.75 & 15.35 & \textbf{23.84} & 26.73\\
LISA+-7B~\cite{yang2025lisaplus} & 37.83 & 46.30 & 39.18 & 44.69 & \phantom{0}8.96 & 12.06 & 31.50\\
MLLMSeg-4B~\cite{mllmseg} & 40.28 & 47.00 & 41.25 & \underline{54.69} & \phantom{0}8.51 & 12.84 & 34.09\\
RemoteReasoner-7B~\cite{yao2026remotereasoner} & 40.39 & 40.71 & 40.61 & 38.12 & 12.14 & 15.24 & 31.20\\
Seg-R1-7B~\cite{you2025segr1} & 41.91 & 42.00 & 45.53 & 43.89 & 13.72 & 18.63 & 34.28\\
VisionReasoner-7B~\cite{liu2026visionreasoner} & 45.11 & 48.30 & \underline{48.59} & 46.54 & \textbf{18.17} & \underline{20.03} & \underline{37.79}\\
Think2Seg-RS-7B~\cite{think2segrs} & \underline{46.40} & \underline{53.07} & 45.28 & 43.99 & 15.13 & 17.82 & 36.95\\
\midrule
GeoScope-4B (Ours) & \textbf{58.18} & \textbf{64.05} & \textbf{62.54} & \textbf{64.82} & \underline{16.39} & 15.30 & \textbf{46.88}\\
\bottomrule
\end{tabular*}
\end{table}

\subsection{Ablation Study}
\label{sec:controlled}

We conduct controlled ablations on \bench{} to verify whether performance gains stem from our scope-aware design rather than output formatting or backbone capacity, and to isolate the role of each module.

\begin{figure}[t]
\centering
\includegraphics[width=\columnwidth]{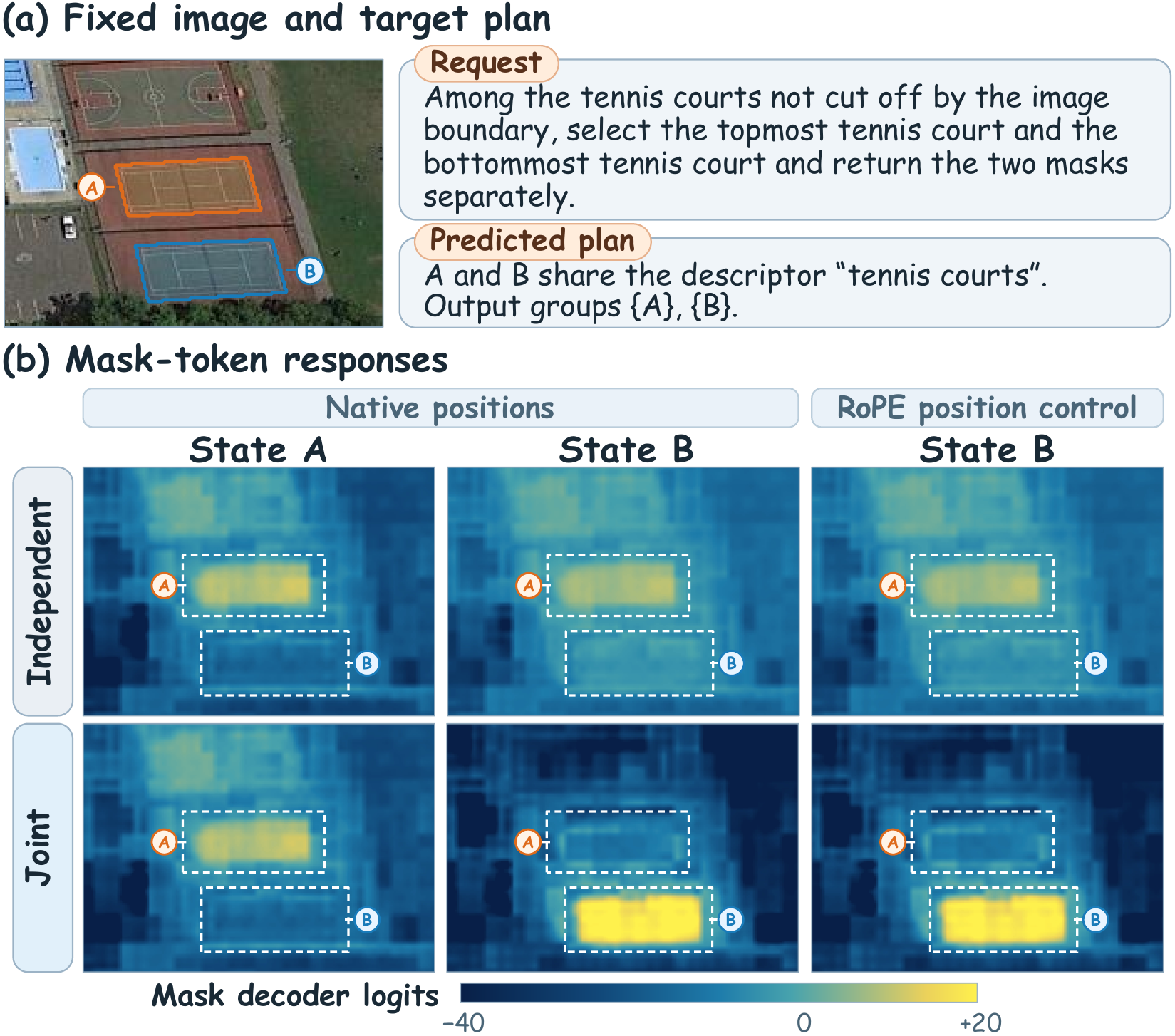}
\caption{\textbf{Mask responses for two targets with the same descriptor.} (a) The predicted plan assigns the courts to separate groups. (b) Both independent states respond to court A. Joint readout shifts the second response to court B at both rotary positions. The heatmaps share a logit scale.}
\label{fig:binding_failure}
\end{figure}

\begin{figure}[!t]
\centering
\includegraphics[width=\columnwidth]{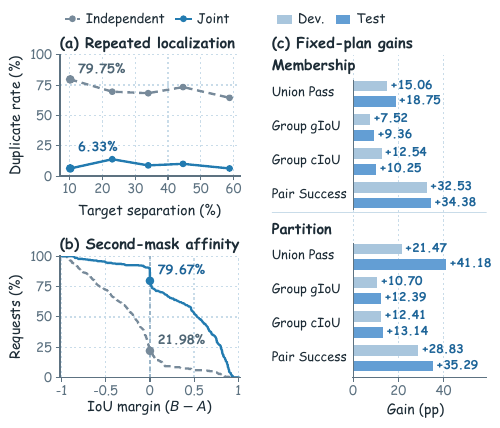}
\captionsetup{skip=4pt}
\caption{\textbf{Target binding under fixed plans.} (a) Duplicate rate by target separation. (b) IoU of the second mask with B minus its IoU with A, where first masks favor A. (c) Gains in union pass rate, group IoU, and Pair Success on Membership and Partition.}
\label{fig:matched_effects}
\end{figure}

\noindent\textbf{Modular Planning across Baseline Paradigms.} To eliminate any disadvantage baselines in Table~\ref{tab:bench} face from unadapted output formatting or grouping rules, Table~\ref{tab:baseline_fairness} directly shares our \bench{}-trained planner and group assembly (\textit{+ Plan \& Group}) with three distinct paradigms: multi-token GSVA-7B~\cite{gsva}, lightweight decoder MLLMSeg-4B~\cite{mllmseg}, and box/point-prompted SAM2 Think2Seg-RS-3B~\cite{think2segrs}. Sharing our in-domain planner consistently improves multi-target execution across all architectures, lifting Partition gIoU by 12--14 points on MLLMSeg-4B and Think2Seg-RS-3B. Yet because these segmenters decode each planned target in isolation without inter-target conditioning, they still duplicate similar instances and trail \method{}.

\begin{table}[t]
\centering\footnotesize
\captionsetup{skip=4pt}
\caption{\textbf{Sharing our trained planner with external baselines on \bench{} test (\%).} \textit{+ Plan \& Group} pairs our planner and group assembly with baseline segmenters (Group gIoU).}
\label{tab:baseline_fairness}
\setlength{\tabcolsep}{2.6pt}
\renewcommand{\arraystretch}{0.95}
\begin{tabular*}{\columnwidth}{@{\extracolsep{\fill}}llccc@{}}
\toprule
Baseline & Configuration & Membership & Partition & Overall\\
\midrule
\multirow{2}{*}{GSVA-7B~\cite{gsva}} & Native & 12.42 & 11.01 & 15.66\\
 & + Plan \& Group & \textbf{14.85} & \textbf{13.97} & \textbf{15.95}\\
\midrule
\multirow{2}{*}{MLLMSeg-4B~\cite{mllmseg}} & Native & 26.50 & 25.35 & 38.46\\
 & + Plan \& Group & \textbf{34.59} & \textbf{38.02} & \textbf{41.96}\\
\midrule
\multirow{2}{*}{Think2Seg-RS-3B~\cite{think2segrs}} & Native & 36.77 & 34.31 & 41.19\\
 & + Plan \& Group & \textbf{42.17} & \textbf{48.30} & \textbf{48.47}\\
\bottomrule
\end{tabular*}
\end{table}

\noindent\textbf{Component Ablation.} Table~\ref{tab:qwen_ablation} isolates our modules under the same Qwen3.5-4B backbone and training data. The vanilla baseline ($\times,\times$) attains 57.08\% overall Group gIoU because single-target slices (Identity, Extent, Paraphrase) need no multi-mask decomposition, yet it collapses on multi-target additions (3.13\% Membership Pair Success). Adding planning ($\checkmark,\times$) lifts Membership gIoU by +6.83 points, but its overall gain is muted because independent readout still produces collinear queries that collapse onto the same instance (6.25\% Pair Success). Coupling planning with Joint Readout ($\checkmark,\checkmark$) conditions later tokens on earlier target states, surging Membership Pair Success to 40.63\% and overall Group gIoU to 61.85\%.

\begin{table}[t]
\centering\footnotesize
\captionsetup{skip=4pt}
\caption{\textbf{Effect of structured planning and joint readout on \bench{} test (\%).} All variants share the identical Qwen3.5-4B backbone, visual features, and in-domain training data.}
\label{tab:qwen_ablation}
\setlength{\tabcolsep}{4.5pt}
\renewcommand{\arraystretch}{0.96}
\begin{tabular*}{\columnwidth}{@{\extracolsep{\fill}}cccccc@{}}
\toprule
\multirow{2}{*}{Plan.} & \multirow{2}{*}{Joint.} & \multicolumn{3}{c}{Membership} & Overall\\
\cmidrule(lr){3-5}\cmidrule(l){6-6}
 & & gIoU & cIoU & Pair & gIoU\\
\midrule
$\times$ & $\times$ & 41.96 & 35.17 & \phantom{0}3.13 & 57.08\\
$\checkmark$ & $\times$ & 48.79 & 36.45 & \phantom{0}6.25 & 57.26\\
\textbf{$\checkmark$} & \textbf{$\checkmark$} & \textbf{58.67} & \textbf{46.70} & \textbf{40.63} & \textbf{61.85}\\
\bottomrule
\end{tabular*}
\end{table}

\noindent\textbf{Earlier Target and Request Context.} In Figure~\ref{fig:binding_failure}, independent readout localizes court A twice, whereas joint readout shifts the second state to court B even under aligned rotary positions~\cite{su2024roformer}; reversing segment order lowers Pair Success after index remapping. When both first masks favor the same target, the joint second mask binds to the alternate instance far more often (79.67\% vs.\ 21.98\%, Figure~\ref{fig:matched_effects}(b)). On public benchmarks, joint readout also boosts LaSeRS \textit{Multiple} by +2.37/+2.69 gIoU/cIoU, and retaining the full request $q$ lifts EarthReason cIoU by +3.60 (+20.74 on fallbacks; Appendices~\docref{app:readout}{E}--\docref{app:public}{F}).

\noindent\textbf{Planning, Grouping, and Error Attribution.} Table~\ref{tab:diagnostics} disentangles upstream planning errors from rendering under a shared joint renderer. Replacing predicted group indices with ground-truth groupings on count-matched predictions (\textit{Annotated groups}) leaves scores unchanged (49.36\% Pair Success), showing that our decoupled planner resolves group topology near-optimally once target count $A$ is correct. Supplying ground-truth target lists (\textit{Annotated plan}) lifts Pair Success to 54.07\% (+12.88 points on Partition), indicating that within the planning stage, instance recall rather than group assignment is the main error source, while remaining failures stem from downstream mask delineation.

\begin{table}[H]
\centering\footnotesize
\renewcommand{\arraystretch}{0.96}
\captionsetup{skip=4pt}
\caption{\textbf{Planning and grouping controls on \bench development (\%).} All conditions use the same joint renderer.}
\label{tab:diagnostics}
\setlength{\tabcolsep}{2.5pt}
\begin{tabular*}{\columnwidth}{@{\extracolsep{\fill}}l*{3}{r}cr@{}}
\toprule
\multirow{2}{*}{Condition} & \multicolumn{2}{c}{Group} & \multicolumn{3}{c}{Pair Success}\\
\cmidrule(lr){2-3}\cmidrule(lr){4-6}
 & gIoU & cIoU & Overall & Membership & Partition\\
\midrule
Predicted plan & 57.75 & 59.84 & 49.36 & 37.95 & 32.52\\
Annotated groups & 57.75 & 59.84 & 49.36 & 37.95 & 32.52\\
Annotated plan & \textbf{59.13} & \textbf{62.00} & \textbf{54.07} & \textbf{44.58} & \textbf{45.40}\\
\bottomrule
\end{tabular*}
\end{table}

\section{Conclusion}
\label{sec:conclusion}
In this work, we investigate scope-aware instruction execution in remote sensing reasoning segmentation. To evaluate this capability, we introduce \bench, which pairs requests on the same aerial image to assess whether masks follow changes in target identity, extent, membership, and grouping. Building on this formulation, we propose \method to specify indexed targets and output groups, jointly read the target states in a causal sequence, and assemble the decoded masks as specified by the plan. Comparisons under fixed plans and a shared renderer show that joint readout improves later-target localization and reduces duplicate predictions. Its gains are most evident when requests add targets or change their grouping. Among the compared methods, \method achieves the best overall results on \bench and the highest mean scores on EarthReason for reasoning segmentation and across three zero-shot referring segmentation benchmarks.

\label{sec:main_end}

\section*{LLM Usage Statement}
During the construction of the dataset, we used the gpt-6.1-sol API to generate instruction pairs. OpenAI Codex helped with the code development and debugging, analysis of experimental results, writing and revision of the paper, as well as the creation of charts. The authors are responsible for the submitted content.

{\small
\bibliographystyle{ieeenat_fullname}
\bibliography{references}
}

\clearpage
\setcounter{figure}{0}
\setcounter{table}{0}
\setcounter{equation}{0}
\renewcommand{\theHfigure}{supp.\arabic{figure}}
\renewcommand{\theHtable}{supp.\arabic{table}}
\renewcommand{\theHequation}{supp.\arabic{equation}}
\maketitlesupplementary
\appendix
\raggedbottom
\setcounter{dbltopnumber}{3}
\renewcommand{\dblfloatpagefraction}{0.76}
\renewcommand{\thefigure}{S\arabic{figure}}
\renewcommand{\thetable}{S\arabic{table}}
\renewcommand{\theequation}{S\arabic{equation}}

\section{Benchmark Construction and Review}
\label{app:construction}

\subsection{Sources and Split Composition}
Development and test splits are strictly disjoint at the source-image level. Table~\ref{tab:data} summarizes the two splits and their paired transformations.

\begin{table}[H]
\centering\small
\captionsetup{skip=4pt}
\caption{\textbf{GeoScope-Bench composition.} Pair counts are grouped by instruction change. Shared instructions are counted once in each split.}
\label{tab:data}
\begin{tabular*}{\columnwidth}{@{\extracolsep{\fill}}lcc@{}}
\toprule
Transformation & Development & Test\\
\midrule
Identity & 167 & 33\\
Extent & 40 & 19\\
Membership & 166 & 32\\
Partition & 163 & 34\\
Paraphrase & 165 & 32\\
\midrule
Pairs & 701 & 150\\
Distinct instructions & 1,392 & 300\\
Images & 345 & 59\\
\bottomrule
\end{tabular*}
\end{table}

\subsection{Annotation and Image Audits}
The programmatic audit enforces six constraints: (1) image and mask dimensions agree; (2) every reference region and output mask is nonempty; (3) the number of output masks matches the declared group count, with every group index valid; (4) the targets in an Identity switch have disjoint masks; (5) the foreground unions in a Partition pair are pixel-identical; and (6) the part mask is a proper subset of its whole-object mask. The frozen release passes all applicable checks across all pairs in the benchmark.

GPT-6.1 Sol~\cite{openai2026gpt61sol} assists candidate query drafting and pre-screens each scene, its paired requests, and the reference overlays for potential ambiguity in target eligibility, count, extent, and grouping. Human reviewers independently verify flagged scenes against the source images, reference masks, and requested scopes.

Split construction prevents source-image overlap between development and test, and an exact-match audit confirms zero decoded-image overlap against the training image pool at the recorded resolution.

\section{Evaluation Metrics}
\label{app:metrics}

We maximize summed IoU with Hungarian assignment~\cite{kuhn1955hungarian}. For a request with $n$ reference and $m$ predicted groups, group gIoU divides the sum of matched IoUs by $\max(n,m,1)$. The dataset score averages over distinct requests. Group cIoU pools intersections and unions across matched groups and requests. Each unmatched mask contributes its area to the pooled union. Union gIoU and cIoU apply the corresponding per-request averaging and area pooling to the two foreground unions.

For an assigned prediction $P$, intended mask $M$ and competing reference $R$, the identity check requires
\begin{equation}
 \begin{aligned}
 &\operatorname{IoU}(P,M)-\operatorname{IoU}(P,R)\\
 &\qquad\geq \min\!\left(0.05,\tfrac{1}{2}[1-\operatorname{IoU}(M,R)]\right).
 \end{aligned}
\end{equation}

\begin{figure}[tbp]
\centering
\includegraphics[width=\columnwidth]{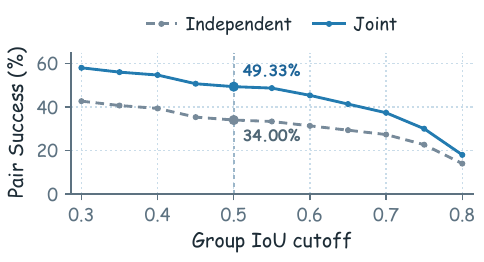}
\captionsetup{skip=4pt}
\caption{\textbf{Pair Success over group IoU cutoffs.} We rescore the test split across cutoffs from $0.30$ to $0.80$ while fixing the other pass criteria. Joint readout preserves a consistent margin over independent readout across lenient and strict thresholds.}
\label{fig:pair_threshold_sensitivity}
\end{figure}

For the transformation relation $C_i$, preserved targets and equal foreground unions require $\operatorname{IoU}\ge 0.7$ across endpoints, while part revisions require preserved area ordering with at least $90\%$ of the part prediction contained inside the whole prediction. Figure~\ref{fig:pair_threshold_sensitivity} further sweeps the group IoU cutoff from $0.30$ to $0.80$ while holding the other pass criteria fixed. Joint readout maintains a substantial margin over independent readout across the entire range, persisting even at the lenient $0.30$ cutoff where boundary jitter is largely forgiven. This confirms that independent readout fails primarily from missed or duplicated instances rather than minor boundary inaccuracy.

\section{Implementation Details}
\label{app:implementation}

\subsection{Task Adaptation and Inference}
\noindent\textbf{Adaptation.} The task adapters build on the Qwen3.5-4B~\cite{qwen2026qwen35} backbone with the native mask decoder (main paper, Sec.~\docref{sec:method}{4} and Sec.~\docref{sec:experiments}{5}.1). The task adaptation uses BF16 forward computation, language LoRA rank 64, alpha 128, dropout 0.05 and effective batch 64. We train with AdamW ($\beta=(0.9,0.95)$), zero weight decay, gradient clipping at 1.0, 3\% warmup and cosine decay to 0.01 of the peak rate. Peak rates are $2\times10^{-4}$ for language LoRA and $2\times10^{-5}$ for the native mask decoder. The original visual encoder stays frozen and images use a 262,144--1,048,576 pixel budget. For LaSeRS and EarthReason, we train and evaluate on their official splits.

\noindent\textbf{Target segments.} Before \masktoken{}, the segment contains \textquotedblleft Target $i$: [description]. Original instruction context: [full request]. Render only target $i$: [description].\textquotedblright{} We wrap the text and the mask token in the native phrase and mask delimiters, respectively. Joint readout concatenates the segments with spaces and appends the end-of-sequence token. Independent readout appends the same token to each segment separately. We decode logits on the original image canvas for \bench and keep the ordered-mask scoring for LaSeRS.

\noindent\textbf{Native mask decoding.} The feature branch projects the pre-merge patches and applies patch self-attention and channel reduction. Two transposed convolutions upsample the spatial features before the target queries score the grid by scaled inner products. Bilinear interpolation maps the logits to the decoder canvas, with a further resize when the evaluation requires the original image resolution.

\noindent\textbf{GeoScope-Bench baseline evaluation.} All methods in Table~\ref{tab:bench_full} are evaluated under unified instruction prompts and identical Hungarian group matching. Planner-transfer controls sharing our planner with external baselines are detailed in Sec.~\ref{app:readout}.

\noindent\textbf{EarthReason baseline evaluation.} In Table~\ref{tab:earthreason_image_overlap}, SegEarth-R1 and Think2Seg-RS-3B use their released task checkpoints, while the remaining baselines are locally adapted on the same EarthReason training split using their native supervision and optimization. At inference, we score the first native mask on the original image canvas and use an empty prediction when the interface returns no mask.

\subsection{Computational Cost}
With $A$ planned targets, joint readout processes $\pi(I,q)$ once in one rendering forward. Independent readout recomputes the prefix for each target without caching or batching across targets. We profile the same frozen renderer on six requests per target count, drawn from development for $A=1,2,3$ and test for $A=4$. Each request is warmed once per condition and timed four times in interleaved order. Table~\ref{tab:runtime_scaling} reports mean synchronized latency from preprocessing through mask logits, excluding planning and grouping, alongside maximum PyTorch memory allocation.

\par\smallskip
\noindent\begin{minipage}{\columnwidth}
\centering\footnotesize
\captionsetup{hypcap=false}
\captionsetup{skip=4pt}
\captionof{table}{\textbf{Rendering latency and peak memory by planned target count $A$.} Timings are measured on one NVIDIA RTX PRO 6000 Blackwell GPU with batch size 1. Bold marks the lower forward count and latency.}
\label{tab:runtime_scaling}
\setlength{\tabcolsep}{2pt}
\begin{tabular*}{\columnwidth}{@{\extracolsep{\fill}}c*{6}{r}@{}}
\toprule
\multirow{2}{*}{$A$} & \multicolumn{2}{c}{Forwards} & \multicolumn{2}{c}{Mean (ms)} & \multicolumn{2}{c}{Peak (GiB)}\\
\cmidrule(lr){2-3}\cmidrule(lr){4-5}\cmidrule(l){6-7}
 & Ind. & Joint & Ind. & Joint & Ind. & Joint\\
\midrule
1 & 1 & 1 & 108.2 & 108.1 & 9.535 & 9.535\\
2 & 2 & 1 & 217.8 & \textbf{118.8} & 9.534 & 9.535\\
3 & 3 & 1 & 323.9 & \textbf{108.8} & 9.535 & 9.538\\
4 & 4 & 1 & 432.6 & \textbf{109.1} & 9.535 & 9.549\\
\bottomrule
\end{tabular*}
\end{minipage}
\par\smallskip

With one target, the two readouts have essentially equal latency. With two to four targets, joint readout is $1.83$--$3.96\times$ faster in this profile, with less than $0.16\%$ additional peak allocated memory.

\section{Additional Benchmark Results}
\label{app:results}

\subsection{Foreground and Group-Level Scores}
Table~\ref{tab:bench_full} gives the full scores for the predictions in the main comparison.

\begin{table*}[tp]
\centering\footnotesize
\captionsetup{skip=4pt}
\caption{\textbf{Full GeoScope-Bench results (\%).} Union scores compare foreground unions, Group scores match output masks, and Pair denotes Pair Success. Bold and underlining denote the best and second-best scores, respectively.}
\label{tab:bench_full}
\setlength{\tabcolsep}{3pt}
\renewcommand{\arraystretch}{0.94}
\begin{tabular*}{\textwidth}{@{\extracolsep{\fill}}l*{10}{c}@{}}
\toprule
\multirow{3}{*}{Method} & \multicolumn{5}{c}{Development} & \multicolumn{5}{c}{Test}\\
\cmidrule(lr){2-6}\cmidrule(lr){7-11}
& \multicolumn{2}{c}{Union IoU} & \multicolumn{2}{c}{Group IoU} & \multirow{2}{*}{Pair}
& \multicolumn{2}{c}{Union IoU} & \multicolumn{2}{c}{Group IoU} & \multirow{2}{*}{Pair}\\
\cmidrule(lr){2-3}\cmidrule(lr){4-5}\cmidrule(lr){7-8}\cmidrule(lr){9-10}
& gIoU & cIoU & gIoU & cIoU & & gIoU & cIoU & gIoU & cIoU & \\
\midrule
Think2Seg-RS-3B~\cite{think2segrs} & 44.51 & 39.38 & 41.54 & 36.05 & \underline{22.40} & 46.69 & 42.84 & 41.19 & 38.92 & 14.67\\
MLLMSeg-4B~\cite{mllmseg} & 37.53 & 38.66 & 34.71 & 36.49 & 16.12 & 41.46 & \underline{51.97} & 38.46 & \underline{50.05} & \underline{18.67}\\
GSVA-7B~\cite{gsva} & 16.37 & 12.61 & 13.77 & 11.66 & 1.57 & 18.48 & 16.36 & 15.66 & 15.34 & 2.00\\
GeoPixel-7B~\cite{geopixel} & 25.64 & 25.41 & 16.70 & 18.52 & 2.71 & 27.29 & 29.65 & 15.20 & 21.72 & 1.33\\
PixelLM-7B~\cite{pixellm} & 25.48 & 19.43 & 22.48 & 17.94 & 7.85 & 31.34 & 27.28 & 27.27 & 24.81 & 11.33\\
LISA+-7B~\cite{yang2025lisaplus} & 29.37 & 24.28 & 28.19 & 23.06 & 10.70 & 33.66 & 35.17 & 32.03 & 33.44 & 10.00\\
PSALM~\cite{zhang2025psalm} & 15.41 & 6.23 & 13.59 & 5.86 & 2.71 & 15.69 & 8.25 & 14.07 & 7.87 & 3.33\\
SegEarth-R1~\cite{segearthr1} & 20.89 & 14.24 & 18.07 & 13.13 & 5.99 & 24.48 & 24.04 & 20.44 & 22.37 & 4.67\\
RemoteReasoner-7B~\cite{yao2026remotereasoner} & 41.22 & 33.91 & 35.92 & 30.69 & 16.41 & 45.10 & 47.37 & 37.98 & 42.27 & 17.33\\
Seg-R1-7B~\cite{you2025segr1} & 43.64 & 32.74 & 37.67 & 28.78 & 18.26 & 44.69 & 33.54 & 38.59 & 29.77 & 16.67\\
VisionReasoner-7B~\cite{liu2026visionreasoner} & \underline{50.23} & \underline{47.87} & \underline{43.47} & \underline{42.86} & 21.97 & \underline{52.73} & 49.89 & \underline{44.50} & 44.49 & 16.00\\
\midrule
GeoScope-4B (Ours) & \textbf{57.80} & \textbf{60.87} & \textbf{58.20} & \textbf{60.27} & \textbf{50.21} & \textbf{62.36} & \textbf{61.77} & \textbf{61.85} & \textbf{60.46} & \textbf{49.33}\\
\bottomrule
\end{tabular*}
\end{table*}

\subsection{Review Sensitivity}
After excluding pairs flagged during pre-screening, GeoScope-4B retains the highest Group IoU and Pair Success among the compared methods on both splits. Pair Success reaches 56.16\% on development and 48.09\% on test.

\subsection{Visualization Examples}
\begin{figure*}[!tp]
\centering
\includegraphics[width=0.95\textwidth]{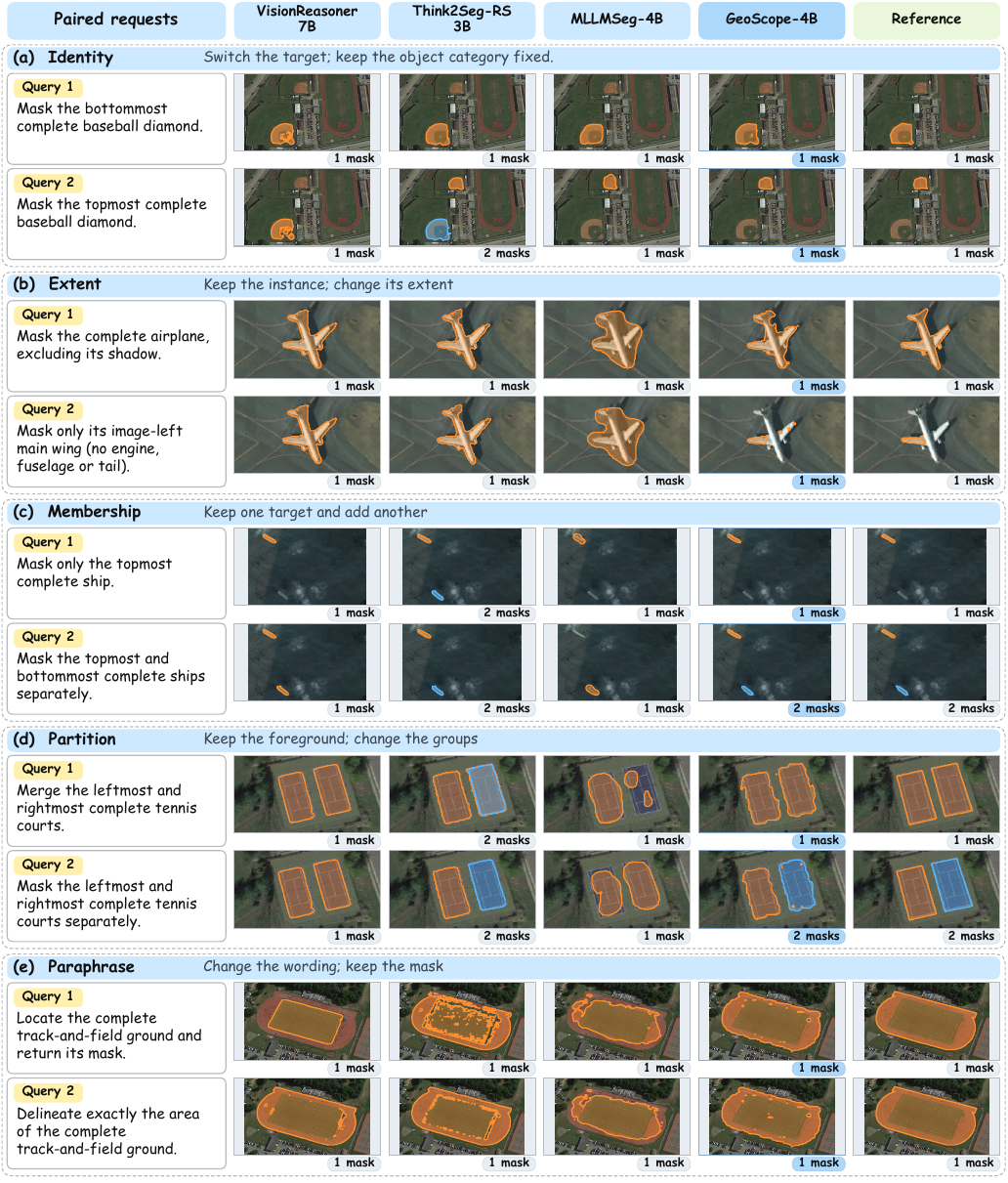}
\captionsetup{skip=4pt}
\caption{\textbf{Qualitative comparisons across all five request transformations.} Each pair uses one development image. Colors distinguish masks within a panel, and the badge gives the mask count. Requests are shortened for display.}
\label{fig:supp_transitions}
\end{figure*}

Figure~\ref{fig:supp_transitions} extends the Membership and Partition comparisons in main-paper Figure~\docref{fig:qualitative}{4} to all five transformations. VisionReasoner-7B~\cite{liu2026visionreasoner}, Think2Seg-RS-3B~\cite{think2segrs}, and MLLMSeg-4B~\cite{mllmseg} use their native interfaces. In Identity, the request switches from the bottommost complete baseball diamond to the topmost one. VisionReasoner still selects the bottommost diamond, while GeoScope follows the new spatial qualifier. In Extent, whole-object predictions persist after the request narrows to a wing. In Membership, VisionReasoner merges the two ships into one mask when the request asks for separate outputs. GeoScope retains the first ship and returns a separate mask for the added ship. In Partition, VisionReasoner covers both courts in one mask after the request specifies separate outputs, while Think2Seg-RS splits them even when one merged mask is requested.

\begin{figure*}[!tp]
\centering
\includegraphics[width=\textwidth]{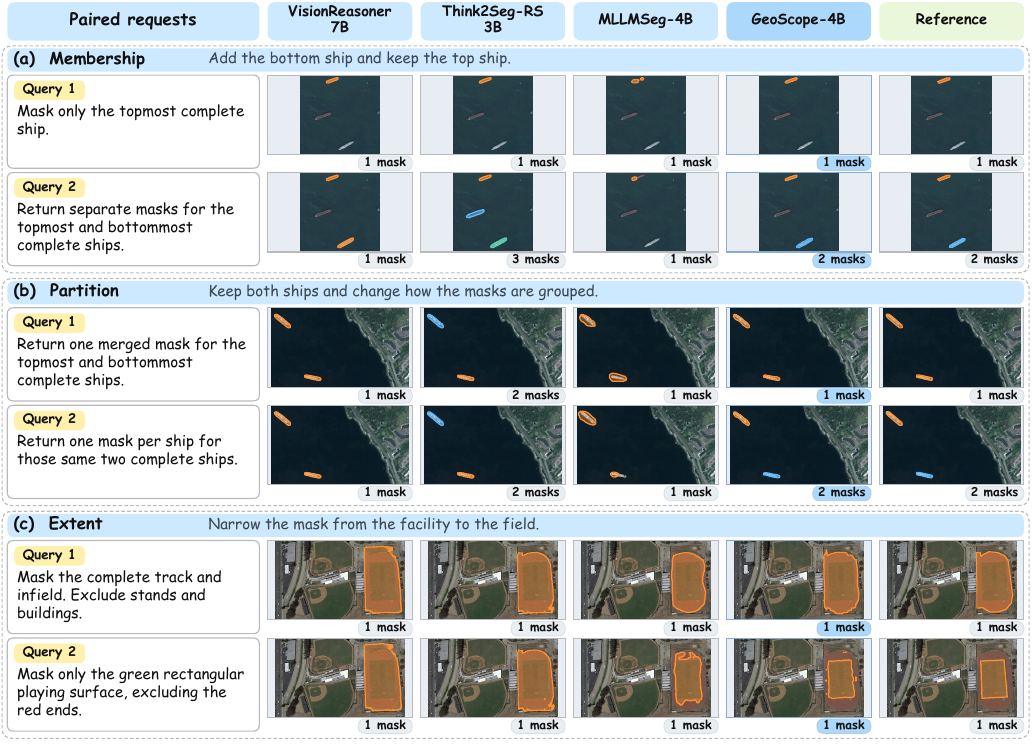}
\captionsetup{skip=4pt}
\caption{\textbf{Qualitative comparisons on the GeoScope-Bench test split.} Saved predictions illustrate target addition, regrouping, and whole-to-part localization. Colors distinguish masks within each panel. Requests are shortened for display.}
\label{fig:supp_test_examples}
\end{figure*}

\noindent\textbf{Test examples.} Figure~\ref{fig:supp_test_examples} shows test pairs that \method passes. The Membership pair retains the top ship and adds the bottom ship as a second mask. In the Partition pair, the two ships remain fixed while the required output changes from one merged mask to one mask per ship. The Extent pair narrows the mask from the entire track-and-field facility to the playing surface inside the track.

\begingroup
\brokenpenalty=10000
\section{Joint Readout Controls}
\label{app:readout}

\subsection{Matched Protocol and Split Results}
We compare independent and joint readout on both splits using the same predicted plans and renderer. The controls below also hold the target segments and visual features fixed, isolating the effect of processing those segments separately or in one sequence.

\noindent\textbf{Planner transfer to external baselines.} In main-paper Table~\docref{tab:baseline_fairness}{4}, GSVA-7B~\cite{gsva}, MLLMSeg-4B~\cite{mllmseg}, and Think2Seg-RS-3B~\cite{think2segrs} each render the targets proposed by the shared planner separately using their native segmentation interfaces. The resulting target masks are then combined according to the planner's output groups. The native baselines instead receive the original requests directly.

\Needspace{9\baselineskip}
\subsection{Order and RoPE Interventions}
For reversal, we concatenate target segments in reverse order, keep their text and target IDs, and map each output state back to its original index before grouping. For the RoPE controls, we compute the difference between the standard joint and independent position IDs of each later mask token. We add that offset to the isolated segment in the independent condition. For the joint control, we subtract it starting at the tested segment while retaining the prefix and earlier segments. We shift all three rotary position channels and verify exact alignment of the later mask-token indices. The first target uses the original positions throughout. The real token order is preserved by each shift.

\begin{table}[tbp]
\centering\footnotesize
\captionsetup{skip=4pt}
\caption{\textbf{Target order and rotary position controls on development (\%).} Results follow the matched fixed-plan protocol of main-paper Table~\docref{tab:diagnostics}{6}. First and Later use the standard joint assignment; Dup.\,$\downarrow$ and Matched IoU use each readout's predictions. In the last four columns, bold and underlining denote the best and second-best scores, respectively.}
\label{tab:mechanism}
\setlength{\tabcolsep}{1.9pt}
\begin{tabular*}{\columnwidth}{@{\extracolsep{\fill}}l*{5}{c}@{}}
\toprule
Readout & First & Later & Dup.$\downarrow$ & Matched IoU & Pair\\
\midrule
Independent & 49.33 & 17.88 & 70.34 & 33.90 & 34.95\\
Indep.\ + Joint RoPE & 49.33 & 17.42 & 70.10 & 33.71 & 34.95\\
Joint (reversed) & 13.94 & 18.02 & 37.75 & 41.52 & 41.08\\
Joint + Indep.\ RoPE & 49.32 & \underline{51.01} & \underline{10.29} & \underline{50.27} & \underline{48.93}\\
Joint & 49.32 & \textbf{51.46} & \textbf{9.80} & \textbf{50.44} & \textbf{49.36}\\
\bottomrule
\end{tabular*}
\end{table}

Table~\ref{tab:mechanism} reports both localization and Pair Success. At fixed token order, assigning independent rotary positions to the joint sequence retains nearly all of the joint readout gain. Conversely, assigning joint positions to independent segments does not recover the gain. The results show that positional offsets alone do not explain the improvement and support a role for earlier target context in distinguishing instances.

\Needspace{4\baselineskip}
\noindent\textbf{Spatial-response measurement.} We retain the continuous decoder logits for multi-target development requests under the five readout conditions. For requests with equal predicted and annotated target counts, we match the standard joint masks to the reference regions by summed IoU and reuse the assignment for every variant. This comparison covers first targets and later targets. The assignment follows the standard joint outputs, so we also compute two per-readout measures over all requests. Duplicate is the share of requests where at least two predicted masks have the same best-overlapping reference with positive overlap. Matched IoU is the mean one-to-one target IoU computed for each condition, with zero for unmatched references (Table~\ref{tab:mechanism}).

For Figure~\docref{fig:matched_effects}{6}(a), we take the requests with two predicted and two reference targets. We divide the reference centroid distance by the image diagonal, sort the requests into five equal-count bins, and plot the duplicate rate at each bin's median distance. For panel (b), both first masks positively overlap the same best-matching reference $A$. We use this shared reference to compute the second mask's IoU margin with the other reference $B$ and plot the fraction of requests above each margin value.

\subsection{Annotation-Assisted Planning and Grouping}
The two interventions in main-paper Table~\docref{tab:diagnostics}{6} use the same joint renderer. For \emph{Annotated groups}, we keep the saved target masks, match them to the annotated regions by maximum summed IoU, and apply the reference groups when the target counts match and both plans cover every target. In that compatible subset, the predicted and annotated partitions already agree after matching, so the intervention leaves the output masks unchanged. For \emph{Annotated plan}, the renderer instead receives the reference target descriptions, count, and groups. This intervention changes the target list available to the renderer as well as the grouping, with the larger improvement on Partition consistent with the need to localize all requested instances before assembling their masks.

\subsection{Request Context Ablation}
The request-context ablation compares the target phrase with the full request $q$, holding target segments, visual features, and independent readout fixed. On GeoScope-Bench development, retaining $q$ raises Group gIoU/cIoU from 41.85/34.15 to 53.28/53.70 and Pair Success from 23.68\% to 35.95\%.

\endgroup

\section{Additional Public Comparisons}
\label{app:public}

\subsection{EarthReason Readout and Image Overlap}
\noindent\textbf{Matched request-context control.} Table~\ref{tab:earthreason_context_full} compares target-phrase and full-request readout on EarthReason~\cite{segearthr1}. The adapted model generates one response per request, from which we extract a shared target phrase. If extraction fails (\emph{Fallback}), both readouts use \textit{the requested region} as the shared generic description. Only the supplied request context changes between the two conditions, explaining the large gain on fallback cases where the generic phrase lacks target-specific cues.

\begin{table}[H]
\centering\footnotesize
\captionsetup{skip=4pt}
\caption{\textbf{Matched EarthReason request context on test (\%).} Parsed uses an extracted target phrase; Fallback uses the shared generic description. Bold denotes the higher score in each row.}
\label{tab:earthreason_context_full}
\setlength{\tabcolsep}{2.4pt}
\begin{tabular*}{\columnwidth}{@{\extracolsep{\fill}}l*{4}{r}@{}}
\toprule
\multirow{2}{*}{Subset} &
\multicolumn{2}{c}{Target phrase} & \multicolumn{2}{c}{Full request}\\
\cmidrule(lr){2-3}\cmidrule(lr){4-5}
& gIoU & cIoU & gIoU & cIoU\\
\midrule
All & 76.49 & 72.84 & \textbf{77.49} & \textbf{76.44}\\
Parsed & 77.15 & 74.99 & \textbf{77.27} & \textbf{75.46}\\
Fallback & 70.34 & 61.16 & \textbf{79.47} & \textbf{81.90}\\
\bottomrule
\end{tabular*}
\end{table}

A small subset of validation and test questions use images whose file bytes match a training image, though their instructions and reference masks are distinct. After excluding those questions, \method reaches 77.18/78.81 gIoU/cIoU on validation and 77.99/80.55 on test (Table~\ref{tab:earthreason_image_overlap}).

\begin{table}[H]
\centering\footnotesize
\captionsetup{skip=4pt}
\caption{\textbf{EarthReason results with image overlap removed (\%).} Bold and underlining denote the best and second-best scores, respectively.}
\label{tab:earthreason_image_overlap}
\setlength{\tabcolsep}{3pt}
\renewcommand{\arraystretch}{0.96}
\begin{tabular*}{\columnwidth}{@{\extracolsep{\fill}}l*{4}{r}@{}}\toprule
&\multicolumn{2}{c}{Validation}&\multicolumn{2}{c}{Test}\\
\cmidrule(lr){2-3}\cmidrule(lr){4-5}
Method & gIoU & cIoU & gIoU & cIoU\\\midrule
GSVA-7B~\cite{gsva} & 51.74 & 47.99 & 50.49 & 48.48\\
LISA+-7B~\cite{yang2025lisaplus} & \underline{71.59} & \underline{71.33} & \underline{72.77} & \underline{73.22}\\
PixelLM-7B~\cite{pixellm} & 65.38 & 61.23 & 65.46 & 62.37\\
PSALM~\cite{zhang2025psalm} & 69.19 & 57.49 & 70.82 & 64.68\\
MLLMSeg-4B~\cite{mllmseg} & 66.71 & 69.43 & 66.82 & 68.93\\
SegEarth-R1~\cite{segearthr1} & 71.05 & 62.89 & 70.75 & 65.22\\
Think2Seg-RS-3B~\cite{think2segrs} & 66.88 & 69.68 & 69.42 & 69.67\\
\midrule
GeoScope-4B (Ours) & \textbf{77.18} & \textbf{78.81} & \textbf{77.99} & \textbf{80.55}\\
\bottomrule
\end{tabular*}
\end{table}

\subsection{LaSeRS}
Table~\ref{tab:lasers_readout_full} extends the fixed-plan readout control in main-paper Sec.~\docref{sec:controlled}{5.4} to LaSeRS~\cite{segearthr2}. The two readouts agree pixel for pixel whenever the shared plan specifies a single target, including every request in the Single and Part slices.
\begin{table}[H]
\centering\footnotesize
\captionsetup{skip=4pt}
\caption{\textbf{Full fixed-plan readout ablation on LaSeRS.} We use the official ordered-mask protocol. Entries are gIoU/cIoU (\%). Avg.\ weights the nine slices equally. Bold highlights slices with measurable readout gains.}
\label{tab:lasers_readout_full}
\setlength{\tabcolsep}{3.2pt}
\renewcommand{\arraystretch}{1.02}
\begin{tabular*}{\columnwidth}{@{\extracolsep{\fill}}lcc@{}}
\toprule
Slice & Independent & Joint\\
\midrule
Semantic & 73.78/80.71 & 73.78/80.70\\
Instance & 78.50/76.17 & 78.50/76.17\\
Part & 77.09/76.59 & 77.09/76.59\\
Single & 68.44/69.13 & 68.44/69.13\\
Multiple & 53.25/71.41 & \textbf{55.61/74.10}\\
Explicit & 86.95/91.10 & 86.95/91.10\\
Implicit & 65.89/80.44 & \textbf{66.13/80.56}\\
Short & 77.54/86.42 & 77.54/86.42\\
Long & 76.12/80.63 & \textbf{76.72/85.37}\\
\midrule
Avg. & 73.06/79.18 & \textbf{73.42/80.02}\\
\bottomrule
\end{tabular*}
\end{table}

\subsection{Qualitative Results on Public Benchmarks}
\label{app:public_visuals}
\begin{figure*}[!tp]
\centering
\includegraphics[width=0.96\textwidth]{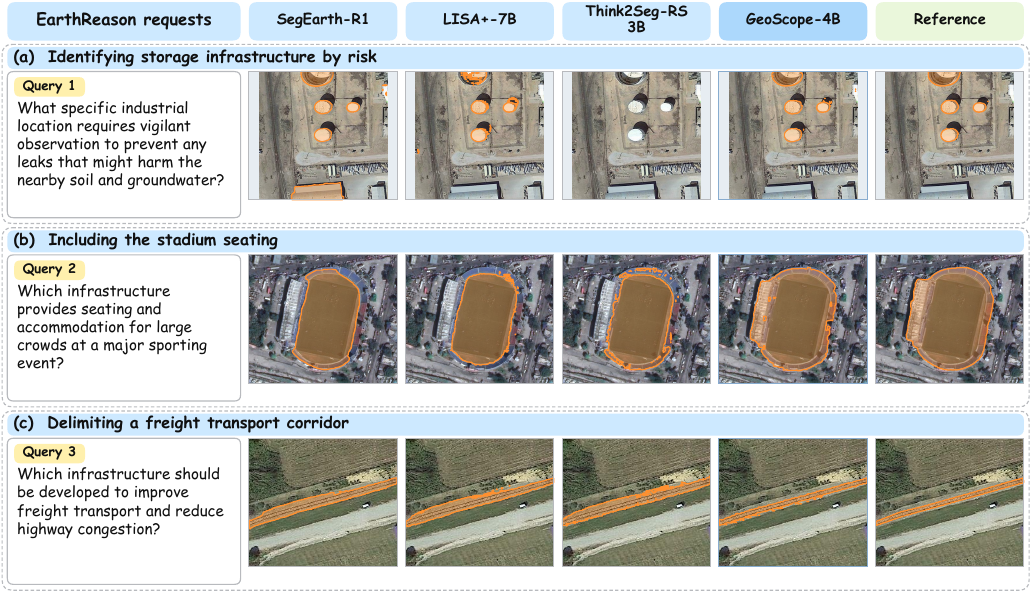}
\captionsetup{skip=4pt}
\caption{\textbf{EarthReason test examples.}}
\label{fig:supp_public_qualitative}
\end{figure*}

\begin{figure*}[!tp]
\centering
\includegraphics[width=0.96\textwidth]{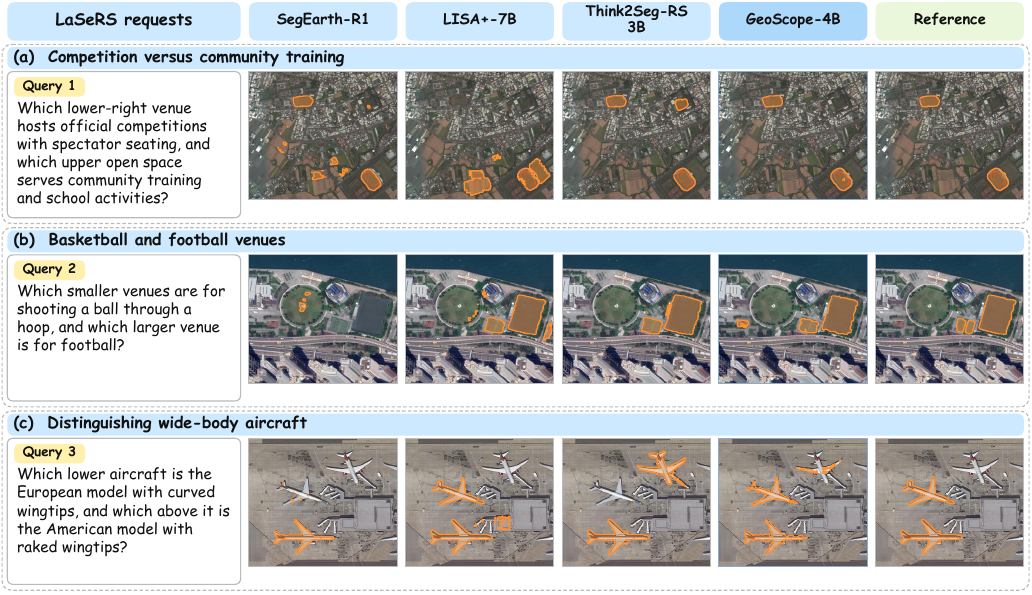}
\captionsetup{skip=4pt}
\caption{\textbf{LaSeRS test examples.}}
\label{fig:supp_lasers}
\end{figure*}

\begin{figure*}[!tp]
\centering
\includegraphics[width=0.96\textwidth]{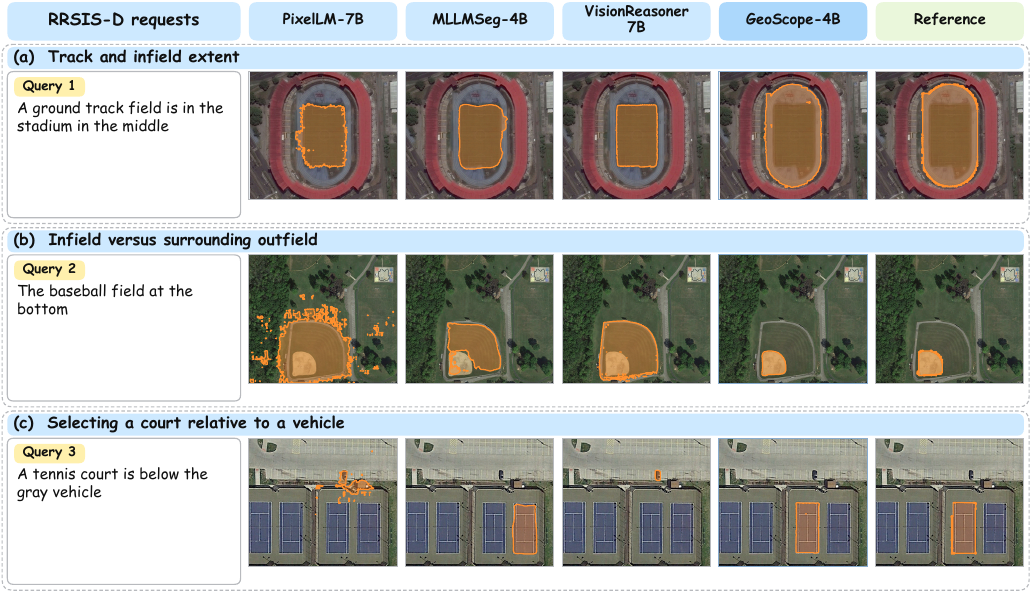}
\captionsetup{skip=4pt}
\caption{\textbf{Zero-shot referring segmentation on RRSIS-D.}}
\label{fig:supp_rrsisd}
\end{figure*}

\begin{figure*}[!tp]
\centering
\includegraphics[width=0.96\textwidth]{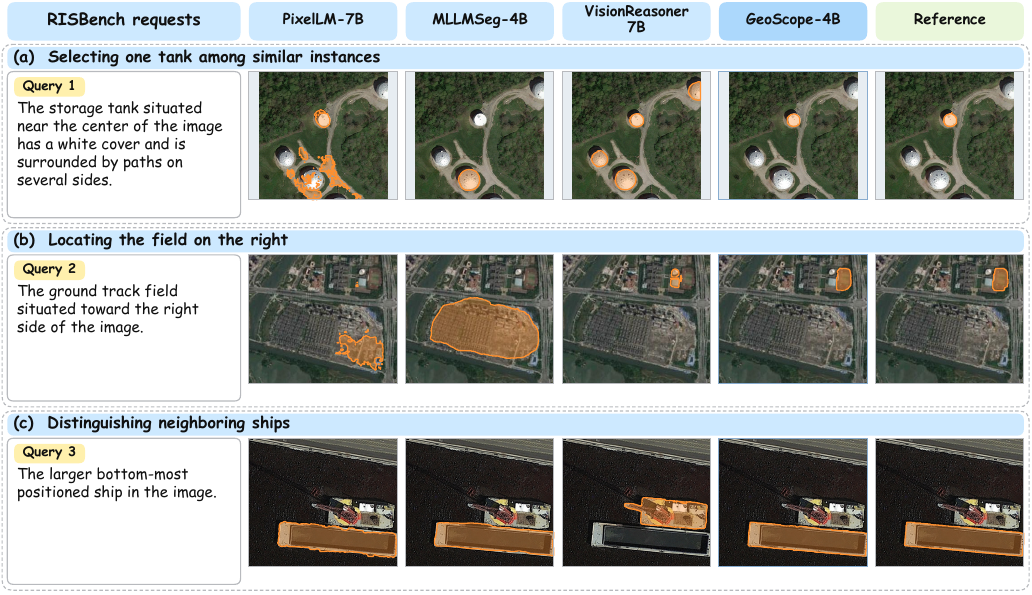}
\captionsetup{skip=4pt}
\caption{\textbf{Zero-shot referring segmentation on RISBench.}}
\label{fig:supp_risbench}
\end{figure*}

\begin{figure*}[!tp]
\centering
\includegraphics[width=0.9\textwidth]{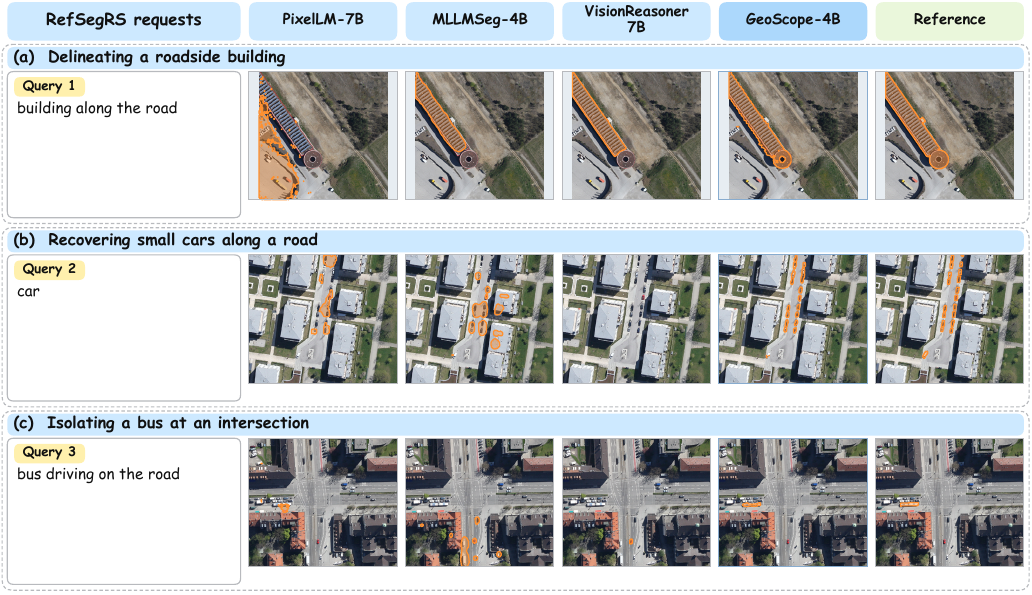}
\captionsetup{skip=4pt}
\caption{\textbf{Zero-shot referring segmentation on RefSegRS.}}
\label{fig:supp_refsegrs}
\end{figure*}

Figures~\ref{fig:supp_public_qualitative}--\ref{fig:supp_refsegrs} show three selected test examples per benchmark, illustrating where GeoScope improves target selection or mask coverage. EarthReason and LaSeRS evaluate reasoning segmentation; we evaluate zero-shot referring segmentation on RRSIS-D~\cite{rmsin}, RISBench~\cite{risbench}, and RefSegRS~\cite{refsegrs}. Each row uses a common view across methods.

\Needspace{16\baselineskip}
\noindent\textbf{EarthReason.} Figure~\ref{fig:supp_public_qualitative} separates errors in target selection from errors in extent. In (a), GeoScope covers the referenced tanks while SegEarth-R1 also includes structures outside the annotated facility. In (b), taking the request about spectator seating into account requires including the grandstand beside the track. In (c), all methods find the freight corridor, but GeoScope more closely follows its narrow annotated boundaries. Aggregate results appear in main-paper Table~\docref{tab:earthreason}{2}.

\noindent\textbf{LaSeRS.} Figure~\ref{fig:supp_lasers} illustrates target selection across requests referring to multiple regions. In (a), GeoScope isolates the competition venue and community training ground, whereas the baselines activate extraneous urban regions. In (b), SegEarth-R1 misses the football pitch entirely, while GeoScope recovers both requested venue types. In (c), SegEarth-R1 and Think2Seg-RS miss the upper-left referenced aircraft, whereas GeoScope captures both referenced wide-body aircraft. These examples complement the fixed-plan readout comparison in Table~\ref{tab:lasers_readout_full}.

\noindent\textbf{RRSIS-D.} Figure~\ref{fig:supp_rrsisd}(a) shows GeoScope extending the mask to the annotated oval boundary, while the baselines cover only the infield. In (b), the reference instead excludes the outfield, and GeoScope follows the smaller extent. In (c), the gray vehicle serves as a landmark: GeoScope selects the court below it, whereas MLLMSeg chooses the neighboring court and VisionReasoner marks the vehicle.

\Needspace{7\baselineskip}
\noindent\textbf{RISBench.} In Figure~\ref{fig:supp_risbench}, GeoScope isolates the white-covered tank near the center in (a) and the track field on the right in (b), avoiding the other instances or structures selected by the baselines. In (c), PixelLM and MLLMSeg also identify the lower ship, while VisionReasoner selects the neighboring ship. GeoScope follows the lower ship's outline more closely.

\noindent\textbf{RefSegRS.} Figure~\ref{fig:supp_refsegrs}(a) shows GeoScope retaining the circular end of the roadside building as part of the requested region. In (b), it covers more of the small cars without labeling nearby roofs. In (c), it localizes the bus at the intersection, whereas the baselines largely miss the target at this scale. The referring galleries complement the aggregate comparisons in main-paper Table~\docref{tab:referring}{3}.

\end{document}